\documentclass[11pt,letterpaper]{article}

\usepackage{csquotes}

\usepackage[title]{appendix}

\usepackage{caption}
\usepackage{subcaption}
\usepackage{fullpage}
\usepackage{times}
\usepackage{array}
\usepackage{ragged2e}
\usepackage{multirow}
\usepackage{fancyhdr, amssymb, mathtools}
\usepackage[ruled,vlined]{algorithm2e}
\usepackage[dvipsnames,table]{xcolor}
\usepackage[margin = 1 in]{geometry}
\usepackage{graphicx}
\usepackage{outlines}
\usepackage{amsmath}

\DeclareMathOperator*{\argmin}{argmin}
\usepackage{graphicx}
\usepackage{dirtytalk}

\usepackage{multirow}
\usepackage{arydshln}

\usepackage[english]{babel} % To obtain English text with the blindtext package
\usepackage{blindtext}
\usepackage{amsfonts}
\usepackage{amssymb}
\usepackage{bbm}
\usepackage{mathtools}

\usepackage{hhline}
\usepackage{makecell}

\usepackage{xltabular}
\usepackage{arydshln} % for dashed lines in table
\usepackage{wrapfig}
\usepackage{enumitem}
\usepackage{floatrow}
\usepackage{xfrac}
\usepackage[font={footnotesize}, labelfont={bf}, labelsep=space, justification=justified, skip=0pt]{caption}
\usepackage[hang,flushmargin]{footmisc} % to remove indentation in footnotes
\usepackage[colorlinks=true, allcolors=blue]{hyperref}

\usepackage{booktabs}
\definecolor{headcolor}{RGB}{255, 255, 255}
\definecolor{columncolor}{RGB}{255, 255, 255}
\definecolor{modelcolor}{RGB}{255, 255, 255}
\definecolor{optioncolor}{RGB}{255, 255, 255}

\definecolor{codegreen}{rgb}{0,0.6,0}
\definecolor{codegray}{rgb}{0.5,0.5,0.5}
\definecolor{codepurple}{rgb}{0.58,0,0.82}
\definecolor{backcolour}{rgb}{0.95,0.95,0.92}

\usepackage{listings}
\lstdefinestyle{mystyle}{
    backgroundcolor=\color{backcolour},   
    commentstyle=\color{codegreen},
    keywordstyle=\color{magenta},
    numberstyle=\tiny\color{codegray},
    stringstyle=\color{codepurple},
    basicstyle=\ttfamily\footnotesize,
    breakatwhitespace=false,         
    breaklines=true,                 
    captionpos=b,                    
    keepspaces=true,                 
    numbers=left,                    
    numbersep=5pt,                  
    showspaces=false,                
    showstringspaces=false,
    showtabs=false,                  
    tabsize=2
}
\usepackage{cleveref} 
\usepackage{cite} % Add the cite package for compressed citation ranges

\AtBeginEnvironment{appendices}{\crefalias{section}{appendix}}

\newcommand\brackets[1]
{\mathopen{}\left[#1\right]\mathclose{}}
\newcommand\parens[1]{\mathopen{}\left(#1\right)\mathclose{}}
\newcommand\braces[1]{\mathopen{}\left\{#1\right\}\mathclose{}}

\newcommand{\cmt}[1]{} % Hides a block of text

\newcommand{\diag}{\text{diag}}
\DeclareMathOperator*{\E}{\mathbb{E}}
\newcommand{\losszeta}{\mathcal{L}(\boldsymbol{\zeta})} % l(\theta), loss function
\newcommand{\losstheta}{\mathcal{L}(\boldsymbol{\theta})} % l(\theta), loss function

\newcommand{\lossthetak}[1]{\mathcal{L}(\boldsymbol{\theta}^{#1})} % L(theta^#1), loss function with #1 as the power

\newcommand{\obj}{l_o}
\newcommand{\Obj}{L_o}

\newcommand{\alphab}{\boldsymbol{\alpha}}

\newcommand{\phib}{\boldsymbol{\phi}}
\newcommand{\rhob}{\boldsymbol{\rho}}

\newcommand{\thetab}{\boldsymbol{\theta}}
\newcommand{\zetab}{\boldsymbol{\zeta}}

\newcommand{\fb}{\boldsymbol{f}}

\newcommand{\hb}{\boldsymbol{h}}

\newcommand{\mb}{\boldsymbol{m}}

\newcommand{\niu}{\mathrm{u}}
\newcommand{\niub}{\boldsymbol{\mathrm{u}}}

\newcommand{\xb}{\boldsymbol{x}}
\newcommand{\Xb}{\boldsymbol{X}}
\newcommand{\nixb}{\boldsymbol{\mathrm{x}}} % non-italic variable
\newcommand{\niXb}{\boldsymbol{\mathrm{X}}} % non-italic variable
\newcommand{\yb}{\boldsymbol{y}}

\newcommand{\niz}{\mathrm{z}} %non-italic z
\newcommand{\nizb}{\boldsymbol{\mathrm{z}}}

\newcommand{\ut}{u_t}
\newcommand{\ux}{u_x}

\newcommand{\uxx}{u_{xx}}

\usepackage{titling}
\thanksheadextra{}{}
\thanksheadextra{}{} % Remove default footnote mark
\usepackage{lipsum} % Example package for filler text

\title{Simultaneous and Meshfree Topology Optimization with Physics-informed Gaussian Processes}
\date{\vspace{-5ex}}
\usepackage{authblk}
\author[1]{Amin Yousefpour}
\author[1]{Shirin Hosseinmardi}
\author[1]{Carlos Mora}
\author[1]{Ramin Bostanabad \thanks{Corresponding Author: Raminb@uci.edu}}
\affil[1]{Department of Mechanical and Aerospace Engineering, University of California, Irvine}

\begin{document}
\include{pythonlisting}
    \pagenumbering{arabic}
    \sloppy
    \maketitle
    \noindent \textbf{Abstract}\\
Topology optimization (TO) provides a principled mathematical approach for optimizing the performance of a structure by designing its material spatial distribution in a pre-defined domain and subject to a set of constraints. The majority of existing TO approaches leverage numerical solvers for design evaluations during the optimization and hence have a nested nature and rely on discretizing the design variables. Contrary to these approaches, herein we develop a new class of TO methods based on the framework of Gaussian processes (GPs) whose mean functions are parameterized via deep neural networks. Specifically, we place GP priors on all design and state variables to represent them via parameterized continuous functions. These GPs share a deep neural network as their mean function but have as many independent kernels as there are state and design variables. We estimate all the parameters of our model in a single for loop that optimizes a penalized version of the performance metric where the penalty terms correspond to the state equations and design constraints. Attractive features of our approach include $(1)$ having a built-in continuation nature since the performance metric is optimized at the same time that the state equations are solved, and $(2)$ being discretization-invariant and accommodating complex domains and topologies. To test our method against conventional TO approaches implemented in commercial software, we evaluate it on four problems involving the minimization of dissipated power in Stokes flow. The results indicate that our approach does not need filtering techniques, has consistent computational costs, and is highly robust against random initializations and problem setup.

    \section{Introduction} \label{sec intro}

Topology optimization (TO) provides a principled mathematical approach for optimizing the performance of a structure by designing its material spatial distribution in a pre-defined domain and subject to a set of constraints. Since the introduction of the homogenization-based approach \cite{RN2023} and its variant SIMP (solid isotropic material with penalization) \cite{RN2025}, TO has become an extremely active research field and a practical design tool \cite{RN2050}. 
The pace of developments in TO is further fueled by the emergence of advanced manufacturing techniques such as additive manufacturing which enable the fabrication of intricate and complex structures that are designed by TO \cite{RN2051}. 

\begin{figure*}[!b]
    \centering
    \includegraphics[width=1.00\columnwidth]{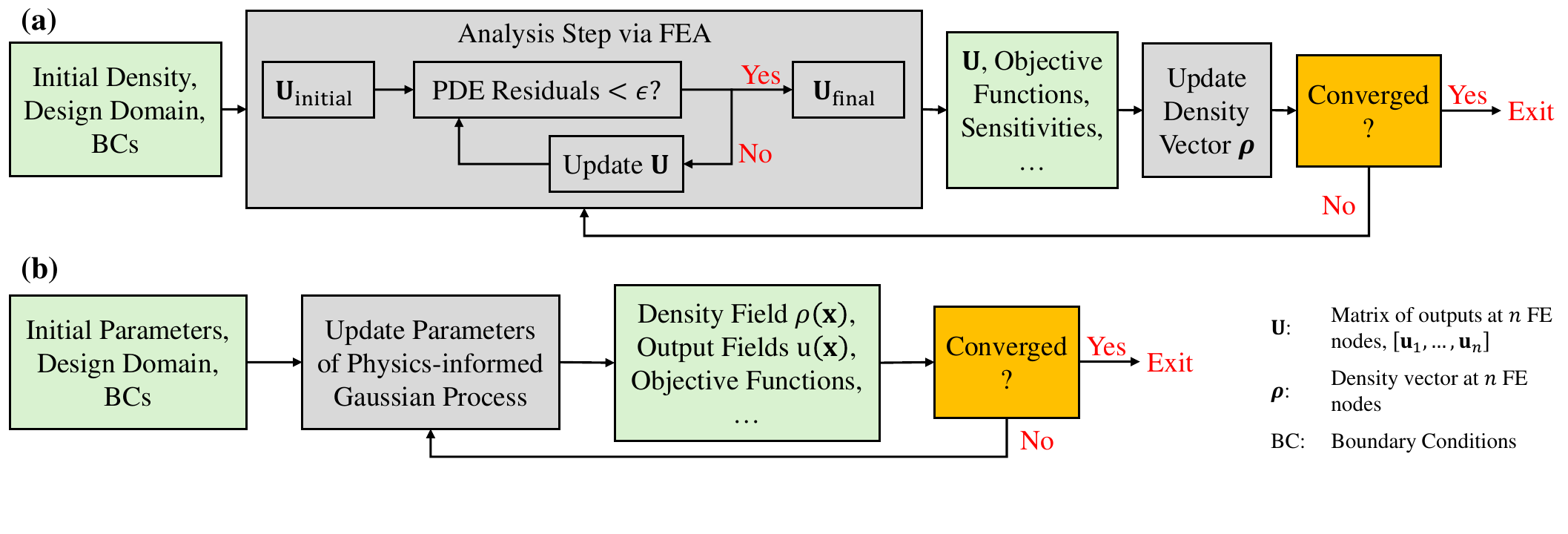}
    \vspace{-1cm}
    \caption{\textbf{Nested and discretized vs simultaneous and mesh-free topology optimization:} Rough flowcharts of the SIMP method \textbf{(a)} and our approach \textbf{(b)} for topology optimization. Algorithmic details are intentionally excluded from these flowcharts.}
    \label{fig generic flowchart}
\end{figure*}

As reviewed in \Cref{sec background}, the majority of existing TO approaches are \textit{nested} \cite
{li2021multidisciplinary,wu2021topology,deaton2014survey,fin2019structural,ben2024robust,stragiotti2024efficient} in that each design iteration involves a nested analysis step where the response of the candidate structure is simulated (typically by solving a system of partial differential equations or PDEs) and then used to update the structure for the next design iteration, see \Cref{fig generic flowchart} for a graphical demonstration. 
To demonstrate this nested process with an example, we consider the TO of fluids in Stokes flow \cite{RN1986}. As detailed in \Cref{sec results}, the goal in this application is to identify the spatial material distribution that $(1)$ is subject to a pre-defined volume constraint, and $(2)$ minimizes the flow's dissipated power. To solve this problem, existing TO methods such as the many variants of SIMP and level-set \cite{RN2021,RN2052} start with an initial design which is then used in an analysis module (e.g., a solver based on the finite element method or FEM) to solve the PDE system that governs the fluid flow in the domain. Once the PDE system is solved, the initial design is updated using the outputs of the analysis module and the analysis-update iterations are continued until a convergence metric is met. In our example on fluids in Stokes flow, these outputs may include the sensitivities (which are needed by SIMP for updating the design variables) or the dissipated power which is the desired performance metric that depends on the fluid flow in the candidate topology. 

%=================================================================
%=================================================================

Another common thread in existing TO methods is that they rely on meshing the structure because the analysis step is typically based on numerical solvers that discretize the design domain. To arrive at mesh-independent designs spatial filters must be used in the optimization loop. Additionally, since the design and state variables are only obtained on the mesh nodes during the optimization, it is common practice to avoid remeshing by treating void as a material with a very small Young's modulus. 
While these choices increase the numerical stability and speed, they rely on manual tuning, especially in applications that involve solid-fluid interaction or large deformations \cite{RN1986}. Meshfree methods have been previously used as solvers in the nested analysis step but these methods are typically more expensive than their mesh-based counterparts and involve more ad-hoc procedures \cite{RN2046, RN2048}. For instance, \cite{RN2042} considers geometrically nonlinear structures and uses the reproducing kernel method to discretize both the displacement and bulk density fields in the governing equations. The design updates are based on SIMP and it is shown that checkerboard issues are observed unless sufficient independent points are used to discretize the density field.

In this work, we develop a \textit{simultaneous} and \textit{meshfree} TO approach that combines the design and analysis steps into a single optimization loop without any nested iterations, see \Cref{fig generic flowchart} for a graphical demonstration. Our approach is founded on physics-informed machine learning (PIML) \cite{mora2024neural,shishehbor2024parametric, RN1920, RN1187, RN1926, RN1886} which has recently emerged as an alternative tool to traditional numerical techniques (such as the FEM) for solving PDEs. 
As briefly reviewed in \Cref{sec pinns review}, the underlying idea behind PIML is to parameterize the solution of a PDE system via a machine learning (ML) model such as a deep neural network (NN) whose loss function and/or architecture is based on the PDE system. Such a setup allows to train the model (and hence approximate the PDE solution) without requiring any labeled solution data inside the domain. Following this idea, herein we develop a new class of ML-based methods for TO where we design the loss function, architecture, and training mechanism of a PIML model that simultaneously optimizes the performance metric (e.g., dissipated power) and satisfies both the design constraints (e.g., solid volume fraction) and the system's governing equations. As detailed in \Cref{sec method}, our PIML model is based on Gaussian processes (GPs) which leverage deep NNs as their mean functions. 

We demonstrate in \Cref{sec results} that our simultaneous TO approach offers a few attractive advantages over existing methods such as SIMP. 
To start with, our method has a built-in continuation nature since the performance metric is optimized at the same time that the system's governing equations are solved. That is, early in the design loop the PDE system (whose solution provides the performance metric) is inexactly and \textit{coarsely} solved since, thanks to the spectral bias of NNs \cite{RN1808}, our PIML model learns the low-frequency part of the PDE solution before its high-frequency components. This built-in continuation nature makes our approach quite insensitive to random initializations while other techniques such as SIMP may show initialization-dependent results and/or computational costs. 

Another advantage of our approach is that it is discretization-invariant and easily accommodates complex domains and topologies. This property is due to the fact that we refrain from meshing the domain. Rather, we parameterize the design and state variables over the domain via a multi-output continuous function whose parameters are optimized by minimizing an appropriately defined loss function that combines the objective function, state equations, and design constraints. As shown in \Cref{sec results}, the discretization-invariant feature of our approach $(1)$ eliminates density filtering requirement (which is needed for SIMP and level-set methods), and $(2)$ reduces the sensitivity of the results to the interpolating functions which are used to convert the discrete design variables to continuous ones that are amenable to gradient-based optimization.

The success of machine learning (ML) in a broad range of critical applications with high visibility has propelled many researchers to gauge the potential of ML in TO. The majority of these developments leverage generative ML models, which are typically built via variational autoencoders (VAEs) \cite{RN941} or generative adversarial networks (GANs) \cite{RN965}, to do instant or iteration-free designs. The essential idea behind many of these approaches is to generate large training data via traditional TO methods such as SIMP and then use that dataset to build a generative ML model that can design a topology with a single forward pass in the network (hence the name instant or iteration-free) \cite{li2019non}. In sharp contrast to these approaches, our TO method does not rely on training data because it is an ML-based solver and not a generative model. 

Besides building generative models, ML has also been used in other capacities such as accelerating the convergence of conventional methods \cite{RN2027, RN2028, RN2029}, parameterizing the density in SIMP \cite{RN1596, RN2044}, or surrogating the FEM-based analysis step \cite{RN2030, RN2031,RN1024,RN1936}. As recently proposed in a review paper \cite{RN1784}, the collection of ML-based works on TO can be broadly categorized into five main groups that include direct design, acceleration, post-processing, reduction, and design diversity. Our approach does not belong to any of these groups as we refrain from coupling ML with existing TO technologies, building generative models that rely on conventional TO methods as data generators, or even designing the architecture and training mechanism of our ML model based on conventional TO methods.

The rest of the paper is organized as follows. 
We review the relevant works in \Cref{sec background} and then introduce our approach in \Cref{sec method} where we provide details on the model's formulations, architecture, loss function, and optimization process. To assess the performance of our PIML-based method, in \Cref{sec results} we study four TO problems on Stokes flow and compare our results to those obtained via the SIMP method implemented in the commercial software COMSOL.
While these four problems only differ in terms of the applied boundary conditions (BCs), their optimum solutions significantly vary. Our comparative studies consider the effects of initialization, overall computational costs, discretization, and randomness. We also analyze the evolution of the topology and the loss function in our approach and compare its accuracy to COMSOL in terms of satisfying the state equations and design constraints. We conclude the paper and provide future research directions in \Cref{sec conclusion}.

\subsection{Nomenclature} \label{subsec nomenclature}
Unless otherwise stated, we denote scalars, vectors, and matrices with regular, bold lower-case, and bold upper-case letters, respectively (e.g., $x, \xb,$ and $\Xb$). Vectors are by default column vectors and subscript or superscript numbers enclosed in parenthesis indicate sample numbers. For instance, $x^{(i)}$ or $\xb^{(i)}$ denote the $i^{th}$ sample in a training dataset while $x_i$ indicates the $i^{th}$ component of the column vector $\xb = \brackets{x_1, \cdots, x_{dx}}^T$. 
% For clarity, we sometimes indicate the size of vectors and matrices via subscripts, e.g., $\xb_{n\times 1}$ and $\Xb_{n\times q}$. 

% We use non-italic fonts to distinguish between a random variable and the specific value that it takes. For instance, $\niy$ and $\niyb$ are a random variable and a vector of random variables, respectively, while $y$ and $\yb$ are a number and a vector of numbers, respectively. 

We distinguish between a function and samples taken from that function by specifying the functional dependence. As an example, $y(x)$ and $y(\xb)$ are functions while $y$ and $\yb$ are a scalar and a vector of values, respectively. 
We also assume functions accommodate batch computations. That is, a single-response function returns a column vector of $n$ values if $n$ inputs are simultaneously fed into it, i.e., $\yb = y(\Xb)$.
When working with multiple functions, we use subscripts to distinguish them. For instance, $y_1(\xb) = sin(x_1) + x_2, y_2(x_1) = x_1^2,$ and $y_3(x_2) = log(x_2^2+1)$ denote three different functions which we may group as $\yb(\xb) = \brackets{y_1(\xb), y_2(x_1), y_3(x_2)}$ where $\xb = \brackets{x_1, x_2}$. 
    \section{Background and Related Works} \label{sec background}
To facilitate the ensuing descriptions, we first formulate a relatively generic TO problem where the goal is to find the spatial material distribution that minimizes the scalar objective function $\Obj(\cdot)$ subject to the set of scalar constraints $C_i(\cdot) \leq 0, i= 1, ..., n_c$.
Denoting the material distribution by the binary spatial density variable $\rho(\nixb)$ where $\nixb = \brackets{x, y, z}$ denotes the spatial coordinates, we can mathematically write this TO problem as:
% ============================================================= %
\begin{subequations}
    \begin{align}
        &\widehat\rho(\nixb) = \underset{\rho\parens{\nixb}}{\argmin} \hspace{2mm}
        \Obj\parens{\niub\parens{\rho(\nixb), \nixb}, \rho(\nixb)} = 
        \underset{\rho\parens{\nixb}}{\argmin} \hspace{2mm}
        \int_\Omega \obj\parens{\niub\parens{\rho(\nixb), \nixb}, \rho(\nixb), \nixb} d\omega, 
        && \forall \nixb \in \Omega 
        \label{eq to generic obj}\\
        & \text{subject to:} \notag \\ 
        &  \qquad \qquad C_1\parens{\rho(\nixb)} = \int_\Omega \rho(\nixb) d\omega - V = 0, && \forall \nixb \in \Omega 
        \label{eq to generic c1} \\
        &  \qquad \qquad C_i\parens{\niub\parens{\rho(\nixb), \nixb}, \rho(\nixb)} = 
        \int_\Omega c_i\parens{\niub\parens{\rho(\nixb), \nixb}, \rho(\nixb), \nixb} d\omega \leq 0, \hspace{2mm} i= 2, ..., n_c, && \forall \nixb \in \Omega 
        \label{eq to generic c2} \\
        &  \qquad \qquad \rho(\nixb) = \braces{0, 1}, && \forall \nixb \in \Omega 
        \label{eq to generic c3}
    \end{align}
    \label{eq to generic}
\end{subequations}
% ============================================================= %
where $\Omega$ denotes the design domain, $V$ is the desired volume fraction\footnote{The equality is replaced via an inequality depending on the application. We write the formulation based on the equality sign as the examples in \Cref{sec results} aim to achieve a desired volume fraction.}, and $\niub(\cdot) = \brackets{\niu_1(\cdot), \cdots, \niu_{n_u}(\cdot)}$ denotes the $n_u$ state variables that must satisfy a set of state equations which typically take the form of a (non-linear) PDE system. In \Cref{eq to generic}, the volume fraction constraint is separated from the rest of the constraints to facilitate the discussions in \Cref{sec method} and it is assumed that the objective function $\Obj\parens{\niub\parens{\rho(\nixb), \nixb}, \rho(\nixb)}$ and constraints $C_i\parens{\niub\parens{\rho(\nixb), \nixb}, \rho(\nixb)}$ can be represented as the integral over appropriately defined local functions\footnote{Due to the integration over the domain, the objective and constraint functions of a given topology (which are scalars) do not depend on $\nixb$ as it is integrated out. However, in \Cref{eq to generic} we explicitly show the dependence of these functions to $\nixb$ because they depend on $\rho(\nixb)$ which varies during the optimization and is indeed a function of $\nixb$.}. 

To solve the TO problem in \Cref{eq to generic}, the popular SIMP method starts by discretizing $\Omega$ via a sufficient number of finite elements and letting $\rho(\nixb)$ take on continuous values in the $\brackets{\rho_{min}, 1}$ range where $\rho_{min}$ is a very small positive number\footnote{With $\rho_{min} = 0$, the domain has to be remeshed as the topology changes during the design iterations.}. That is: 
% ============================================================= %
\begin{subequations}
    \begin{align}
        &\widehat\rhob = \underset{\rhob}{\argmin} \hspace{2mm}
        \Obj\parens{\niub\parens{\rhob}, \rhob} = 
        \underset{\rhob}{\argmin} \hspace{2mm}
        \sum_{i=1}^{n_{\Omega}}\int_{\Omega_i} \obj\parens{\niub\parens{\rho_i}, \rho_i} d\omega_i, 
        && \forall \nixb \in \Omega 
        \label{eq to dis obj}\\
        & \text{subject to:} \notag \\ 
        &  \qquad \qquad C_1\parens{\rhob} = \sum_{i=1}^{n_{\Omega}} \rho_i \omega_i - V = 0, && \forall \nixb \in \Omega 
        \label{eq to dis c1} \\
        &  \qquad \qquad C_i\parens{\niub\parens{\rhob}, \rhob} = 
        \sum_{j=1}^{n_{\Omega}}\int_{\Omega_j} c_i\parens{\niub\parens{\rho_j}, \rho_j} d\omega_j \leq 0, \hspace{2mm} i= 2, ..., n_c, && \forall \nixb \in \Omega 
        \label{eq to dis c2} \\
        &  \qquad \qquad \rho_i \in \brackets{\rho_{min}, 1}, \hspace{2mm} i= 1, ..., n_{\Omega}, && \forall \nixb \in \Omega 
        \label{eq to dis c3}
    \end{align}
    \label{eq to dis generic}
\end{subequations}
% ============================================================= %
where $n_{\Omega}$ is the number of finite elements, $\rhob = [\rho_1, \cdots, \rho_{n_{\Omega}}]^T$ is the vector of element densities with $\rho_i$ being the density of the element at $\nixb_i$, $\omega_i$ is the volume of the $i^{th}$ element, and $\Omega = \cup_{i=1}^{n_\Omega} \Omega_i$. 
In this setting, the original TO problem is cast as estimating the material density in each finite element rather than for any arbitrary point in the domain. To encourage $1$ or $0$ density values an interpolating function with penalization is used. For instance, in the context of minimizing the compliance in structural mechanics, this process involves scaling the material properties via a power-law function. Taking Young's modulus as an example property, power-law scaling converts $E\parens{\nixb_i}$ to $\rho\parens{\nixb_i}^p E_0$ where $p$ is the penalty factor, $\nixb_i$ denotes the $i^{th}$ finite element, and $E_0$ is the Young's modulus of the solid material. In compliance minimization applications $p$ is typically set to $3$ but for the examples in \Cref{sec results} we use $p=2$ as it yields better results. 

The proper choice of the interpolating function depends on the application. This is especially the case in fluid problems such as the ones studied in \Cref{sec results} that involve pressure minimization. In such cases, more sophisticated interpolation functions are needed to encourage binary density values while smoothing the influence of the topology on the flow and preventing premature convergence \cite{RN1986}.

It is well-known that the design space corresponding to \Cref{eq to generic} or its variation in \Cref{eq to dis generic} are not closed in that the obtained topology depends on the element size and mesh resolution. This issue can be simply addressed via restriction methods such as sensitivity filtering where the element sensitivities are spatially averaged within a mesh-independent region \cite{RN2023}. Density filtering \cite{RN2023} is another effective restriction method that spatially averages $\rho\parens{\nixb}$ to obtain $\bar{\rho}\parens{\nixb}$ which is then interpreted as the physical density of an element. The density filter can be formulated based on an inhomogeneous Helmholtz-type equation with Neumann boundary conditions (BCs) \cite{RN2034}:
% ============================================================= %
\begin{subequations}
    \begin{align}
        & -r^2\nabla^2\bar{\rho}\parens{\nixb} + \bar{\rho}\parens{\nixb} = \rho\parens{\nixb}, && \forall \nixb \in \Omega\\
        & \frac{\partial \bar{\rho}\parens{\nixb}}{\partial \boldsymbol{n}} = 0, && \forall \nixb \in \Omega
    \end{align}   
    \label{eq: helmholtz}
\end{subequations}
% ============================================================= %
where $\nabla^2$ denotes the Laplacian operator, $r$ controls the size of the averaging neighborhood, and $\boldsymbol{n}$ denotes the surface normal vector.

A disadvantage of sensitivity and density filters is that they result in topologies with grey transitions where the density is not sufficiently close to $0$ or $1$. To address this issue, projection methods \cite{RN2035, RN2036} are used where the filtered density field $\bar{\rho}\parens{\nixb}$ is passed through a differentiable Heaviside projection function. The combination of a filter and a projection \cite{RN926} is akin to a parameterized level set approach which we briefly review next. 

Originally developed for optimizing linearly elastic structures, the level set approach \cite{RN2024, RN2037, RN2038} implicitly represents the boundaries of a structure by the zero-level of the scalar level set function $\psi(\nixb)$ and the structure is defined as the region over which $\psi(\nixb)$ is positive. That is:
% ============================================================= %
\begin{equation}
    \rho\parens{\nixb} = 
    \begin{cases}
        0, & \psi(\nixb) < 0 \\
        1, & \psi(\nixb) \geq 0 
    \end{cases}
    .
\end{equation}
% ============================================================= %

In this approach, design variations are achieved by updating the implicit representation of the structure boundaries. These updates are typically formulated via the Hamilton-Jacobi equation whose so-called speed function, which controls the sensitivity of moving the $\psi(\nixb) = 0$ interface, is defined based on the objective function. The original Hamilton-Jacobi equation only has a convective term but recent works have augmented it via diffusive and reactive terms to regularize the optimization or enable nucleation of new holes \cite{RN2039}.

While other TO approaches have been developed over the past few decades \cite{RN2040, RN930}, SIMP and level set are the two most popular ones which are both gradient-based and rely on nested analysis steps to update the initial design. As briefly mentioned in \Cref{sec intro} and extensively reviewed in \cite{RN1784}, ML methods have been increasingly used to either replace these methods (but after using them for data generation) or augment them to achieve, e.g., computational efficiency. 

As explained in the next section, the fundamental idea behind our approach differs from most existing works that combine ML with conventional TO methods because we aim to solve \Cref{eq to generic} along with the state equations that govern $\niub(\nixb)$ simultaneously via ML. That is, training our ML model amounts to solving a constrained optimization problem. Recently some works \cite{RN2041, RN2043, RN2045} have explored directions that share some similarities with our method. However, these works are specifically designed for compliance minimization, require parameter tuning across different problems, and still have a nested nature since they are inspired by SIMP and their algorithms involve nested for loops while our method has only one for loop.
For example, Ntopo is proposed in \cite{RN2041} and integrates two deep NNs that parameterize the density and displacement fields as functions of spatial coordinate systems. In each iteration of the design, the density network is fixed and the displacement network is trained to satisfy the equilibrium equation that is based on the potential energy function. The decoupled nature of these two networks simplifies the training but removes structural meaning and necessitates the use of filters. Inspired by Ntopo, \cite{RN2045} leverages the deep energy method (DEM) which is limited to applications where a potential energy function can be formulated. The authors of \cite{RN2045} leverage the fact that compliance minimization is a self-adjoint problem to dispense with training the displacement network and instead obtain the element sensitivities directly from the DEM model.

    \section{Proposed Methodology} \label{sec method}
The starting point for simultaneous and meshfree TO is to include the state equations that govern $\niub(\nixb)$ as additional constraints in \Cref{eq to generic}. For many applications such as the one studied in \Cref{sec results}, these equations take the form of PDE systems. Adding these state constraints to \Cref{eq to generic} and introducing the intermediate continuous density variable, $\widetilde\rho\parens{\nixb}$, we obtain:
% ============================================================= %
\begin{subequations}
    \begin{align}
        &\widehat\rho(\nixb) = \underset{\rho\parens{\nixb}}{\argmin} \hspace{2mm}
        \Obj\parens{\niub\parens{\nixb}, \rho(\nixb)} = 
        \underset{\rho\parens{\nixb}}{\argmin} \hspace{2mm}
        \int_\Omega \obj\parens{\niub\parens{\nixb}, \rho(\nixb), \nixb} d\omega, 
        && \forall \nixb \in \Omega 
        \label{eq sim to generic obj}\\
        & \text{subject to:} \notag \\ 
        &  \qquad  C_1\parens{\rho(\nixb)} = \int_\Omega \rho(\nixb) d\omega - V = 0, && \forall \nixb \in \Omega 
        \label{eq sim to generic c1} \\
        &  \qquad  C_i\parens{\niub\parens{\nixb}, \rho(\nixb)} = 
        \int_\Omega c_i\parens{\niub\parens{\nixb}, \rho(\nixb), \nixb} d\omega \leq 0, \hspace{2mm} i= 2, ..., n_c, && \forall \nixb \in \Omega 
        \label{eq sim to generic c2} \\
        &  \qquad  R_i\parens{\niub\parens{\nixb}} = 
        \int_\Omega r_i\parens{\niub\parens{\nixb}, \nixb} d\omega = 0, \hspace{2mm} i= 1, ..., n_r, && \forall \nixb \in \Omega 
        \label{eq sim to generic c3} \\        
        &  \qquad  \rho(\nixb) = g\parens{\widetilde\rho\parens{\nixb}}, 
        \hspace{2mm} \rho(\nixb) \in \brackets{0, 1}, && \forall \nixb \in \Omega 
        \label{eq sim to generic c4}
    \end{align}
    \label{eq sim to generic}
\end{subequations}
% ============================================================= %
where $g(\cdot)$ is a differentiable projection function that promotes $0/1$ density values and $R_i(\niub(\nixb))$ is the constraint corresponding to the $i^{th}$ state equation or initial and boundary conditions (ICs and BCs). In \Cref{eq sim to generic} it is assumed that $R_i(\niub(\nixb))$ can be calculated as the integral of the local function $r_i(\niub(\nixb), \nixb)$. 
It is noted that compared to \Cref{eq to generic}, we are no longer writing $\niub(\nixb)$ to explicitly depend on $\rho(\nixb)$ in \Cref{eq sim to generic} because our approach is not nested, i.e., we find $\rho(\nixb)$ and $\niub(\nixb)$ jointly rather than designing $\rho(\nixb)$ first and then solving for $\niub(\nixb)$ in an analysis step.

For a structure that is governed by a PDE system, the set of $R_i(\niub(\nixb))$ includes differential equations as well as ICs and BCs. The nested TO methods reviewed in \Cref{sec background} require a candidate topology, optimum or not, to produce zero (or very small) residuals on these constraints. Additionally, the majority of these TO methods rely on meshing the domain by a finite number of elements which 
reduces the dimensionality of the problem by solving for the vector $\rhob = [\rho_1, \cdots, \rho_{n_{\Omega}}]^T$ instead of the function $\rho(\nixb)$.

In our approach, we dispense with the meshing and nesting requirements by directly solving the constrained optimization problem formulated in \Cref{eq sim to generic}. These choices provide a few attractive features. 
First, we do not observe checkerboard patterns or mesh-dependency issues since the design domain is never discretized. Hence, filtering techniques such as the one in \Cref{eq:  helmholtz} are not required to arrive at mesh-invariant solutions.
Second, we do not require the state variables to strictly satisfy their state equations during the entire optimization process. Specifically, at the early stages of the optimization we mainly explore the design space and then focus more on satisfying $R_i(\niub(\nixb))$ as the design iterations increase. This behavior provides our approach with a built-in continuation nature and, as shown in \Cref{sec results}, makes it quite robust to random initializations. 
The third advantage of our approach is that it is end-to-end and does not require nesting an analysis module within an optimization process. 

The major challenge associated with our approach is that it is considerably difficult to solve the constrained optimization problem in \Cref{eq sim to generic} especially since the degree of non-linearity and scale of the objective function and the constraints are generally very different. 
To address this challenge, we begin by parameterizing all the dependent variables, that is $\niub\parens{\nixb}$ and $\rho\parens{\nixb}$, with the differentiable multi-output function $\nizb(\nixb; \zetab)$ whose last output corresponds to $\rho\parens{\nixb}$. That is:
% ============================================================= %
\begin{equation}
    \nizb(\nixb; \zetab) \coloneqq \brackets{\niu_1(\nixb), ..., \niu_{n_{\niu}}(\nixb), \rho\parens{\nixb}},
    \label{eq z definition}
\end{equation}
% ============================================================= %
where $\zetab$ denotes the parameters of $\nizb(\nixb; \zetab)$.
For notational simplicity, we define the following two shorthands: $\niz_{-1}(\nixb; \zetab) := \rho\parens{\nixb}$ and $\nizb_{\sim1}(\nixb; \zetab) := \niub\parens{\nixb}$. We note that not all the elements in $\zetab$ are used for estimating both $\rho(\nixb)$ and $\niub\parens{\nixb}$ but we do not explicitly indicate this in the two shorthands to avoid complicating the notation.

Having defined $\niz_{-1}(\nixb; \zetab)$ and $\nizb_{\sim1}(\nixb; \zetab)$ we now convert \Cref{eq sim to generic} to the unconstrained optimization problem in \Cref{eq penalty opt generic} via the penalty method to estimate $\widehat\zetab$ and, in turn, simply obtain the designed topology via $\niz_{-1}(\nixb; \widehat\zetab)$:
% ============================================================= %
\begin{equation}
    \begin{aligned}
        \widehat\zetab = \underset{\zetab}{\argmin} \hspace{2mm} \losszeta
        &= \underset{\zetab}{\argmin} \hspace{2mm} 
        \Obj\parens{\nizb(\nixb; \zetab)} + 
        \mu_p \left( 
        \sum_{i=1}^{n_r}\alpha_i R_i^2\parens{\nizb_{\sim1}(\nixb; \zetab)} \right. \\
        & \left. + \alpha_{n_r + 1}C_1^2\parens{\niz_{-1}(\nixb; \zetab)} + 
        \sum_{i=2}^{n_c}\alpha_{n_r+i} \max\parens{0, C_i^2\parens{\nizb(\nixb; \zetab)}} \right)
    \end{aligned}
    \label{eq penalty opt generic}
\end{equation}
% ============================================================= %
where $\mu_p$ is the penalty factor and $\alphab = \brackets{\alpha_1, \cdots, \alpha_{n_r+n_c}}$ are weights that ensure none of the terms on the right-hand side of \Cref{eq penalty opt generic} dominates $\losszeta$. As detailed in \Cref{subsec training} we use a first order gradient-based optimizer to solve \Cref{eq penalty opt generic} where we set $\alpha_1 = \alpha_2 = 1$ to use $R_1(\niub\parens{\nixb})$\footnote{$R_1(\niub\parens{\nixb})$ and $R_2(\niub\parens{\nixb})$ represent momentum residuals in our applications and have similar magnitudes. Hence, we set $\alpha_1 = \alpha_2 = 1$.} as a reference for scaling the other terms that penalize $\Obj\parens{\nizb(\nixb; \zetab)}$.
To calculate the individual terms on the right-hand side of \Cref{eq penalty opt generic} we first represent each one as an integral of a local function and then estimate the integration via a summation. That is:
%============================================================ %
\begin{subequations}
    \begin{align}
        & \Obj\parens{ \nizb(\nixb; \zetab)} = \int_{\Omega} \obj\parens{ \nizb(\nixb; \zetab), \nixb} d\omega \approx
        \sum_{i=1}^{n_{CP}} \nizb(\nixb_i; \zetab) \omega_i, 
        \label{eq sim to dis lo} \\    
        & R_i\parens{\nizb_{\sim1}(\nixb_i; \zetab)} = 
        \int_{\Omega} r_i\parens{\nizb_{\sim1}(\nixb; \zetab), \nixb} d\omega \approx
        \sum_{j=1}^{n_{CP}} r_i\parens{\nizb_{\sim1}(\nixb_j; \zetab)} \omega_j
        = 0, \hspace{2mm} i= 1, ..., n_r,
        \label{eq sim to generic ri}\\               
        & C_1\parens{ \niz_{-1}(\nixb; \zetab)} = 
        \int_{\Omega} c_1\parens{ \niz_{-1}(\nixb; \zetab), \nixb} d\omega - V \approx
        \sum_{j=1}^{n_{CP}} \niz_{-1}(\nixb_j; \zetab) \omega_j - V = 0, 
        \label{eq sim to dis c1} \\
        & C_i\parens{\nizb(\nixb; \zetab)} = 
        \int_{\Omega} c_i\parens{\nizb(\nixb; \zetab), \nixb} d\omega \approx
        \sum_{j=1}^{n_{CP}} c_i\parens{\nizb(\nixb_j; \zetab)} \omega_j \approx
        \leq 0, \hspace{2mm} i= 2, ..., n_c,
        \label{eq sim to generic ci}      
        % & \niz_{-1}(\nixb; \zetab) = g\parens{\widetilde\niz_{-1}(\nixb; \zetab)}, 
        % \hspace{2mm} \niz_{-1}(\nixb; \zetab)\in \brackets{0, 1}, 
        % \label{eq sim to generic c3}      
    \end{align}
    \label{eq penalty opt details}
\end{subequations}
% ============================================================= %
where $\nixb_j$ are a set of $n_{CP}$ collocation points (CPs) that are distributed in the design domain. We highlight that the summations in \Cref{eq penalty opt details} aim to approximate the integrals and should not be confused with the summations in \Cref{eq to dis generic} which arise from discretizing the design domain and variables.

We now return to $\nizb(\nixb; \zetab)$ which parameterizes our search space. Since the state variables depend on the topology, it is clear that the representation of $\niub\parens{\nixb}$ and $\rho(\nixb)$ should share (at least some) parameters. In addition, $\nizb(\nixb; \zetab)$ must have sufficiently high representation power to be able to encode a wide range of topologies and state variables; otherwise different functional forms must be devised for $\nizb(\nixb; \zetab)$ in different problems. Lastly, we must be able to efficiently calculate the gradients of $\nizb(\nixb; \zetab)$ with respect to $\zetab$ and $\nixb$\footnote{Gradients with respect to $\nixb$ are typically required for satisfying the state equations that govern $\niub\parens{\nixb}$.} since \Cref{eq penalty opt generic} is best minimized by a gradient-based optimization approach. 

The above requirements motivate us to use an ML model as $\nizb(\nixb; \zetab)$, specifically a GP whose mean function is a deep NN. In addition to satisfying the above requirements, this ML model has the attractive property that it can satisfy the BCs/ICs by construction, i.e., these constraints can be excluded from \Cref{eq penalty opt generic}. In the following subsections, we elaborate on the architecture and properties of our ML model in \Cref{subsec model description} and then explain how we efficiently minimize $\losszeta$ in \Cref{subsec training}.

\subsection{Parameterization of Design Space and State Variables} \label{subsec model description}
The design space is represented by $\rhob(\nixb)$ and parameterized via $\niz_{-1}(\nixb; \zetab)$. Unlike $\rhob(\nixb)$ which only has to be in the $[0, 1]$ interval (and ideally take on values that are very close to either $0$ or $1$), the state variables $\niub\parens{\nixb}$ must satisfy the PDE system that governs the structure's response. To elaborate on how this requirement affects our parameterization, we consider the following generic boundary value problem\footnote{The descriptions naturally extend to time-dependent PDE systems, for details see \Cref{sec pinns review} and \cite{mora2024neural}.}:
%=================================================================
\begin{subequations}
    \begin{align}
        &\mathcal{N}_{\nixb}[\niub(\nixb)] = \fb(\nixb), \quad \nixb \in \Omega,
        \label{eq generic pde}\\
        &\niub(\nixb) = \hb(\nixb), \quad \quad \nixb \in \partial \Omega,
        \label{eq generic bc}
    \end{align}
    \label{eq generic pde system}
\end{subequations}
%=================================================================
where \( \partial\Omega \) denotes the boundary of $\Omega$, $\mathcal{N}_{\nixb}[\niub(\nixb)]$ are a set of differential operators acting on $\niub(\nixb)$, \( \fb(\nixb) = \brackets{f_1(\nixb), \cdots, f_{n_u}(\nixb)} \) is a known vector of functions, and the prescribed BCs on the state variables are characterized via \( \hb(\nixb) = \brackets{h_1(\xb), \cdots, h_{n_u}(\xb)} \). 

\Cref{eq generic bc,eq generic pde} can both be converted to equality constraints and inserted in \Cref{eq penalty opt generic}. In practice, however, there are two major issues with these conversions. 
First, \Cref{eq generic bc} depends on $\niub\parens{\nixb}$ while \Cref{eq generic pde} is formulated based on the gradients of $\niub\parens{\nixb}$. As a result, the scales of the corresponding constraints in \Cref{eq penalty opt generic} are substantially different which requires a careful adjustment of the weights $\alphab$. 
Second, the effectiveness of the previously mentioned continuation effect significantly decreases if it involves both BCs and the differential equations. This adverse effect manifests itself by increasing the optimization iterations required to minimize \Cref{eq penalty opt generic}; increasing the overall computational costs and perhaps negating the advantages of continuation. 
We attribute this behavior to the fact that the solution of a PDE system strongly depends on the applied BCs and an estimate solution that does not strictly satisfy them results in a largely erroneous search direction while minimizing $\losszeta$.

Motivated by our recent work \cite{mora2024neural}, we address the above two issues by designing an ML model that strictly satisfies the BCs/ICs imposed on $\niub\parens{\nixb}$ which, in turn, eliminates the need to incorporate \Cref{eq generic bc} as a constraint into \Cref{eq penalty opt generic}.
Specifically, we first put a GP prior on each of the $n_u$ state variables. As justified below, we endow these GPs with independent kernels $k_i(\nixb, \nixb'; \phib_i)$ but a single shared multi-output mean function that is parameterized by a deep NN, i.e., $\mb(\nixb; \thetab) = \brackets{m_1(\nixb; \thetab), \cdots, m_{n_u}(\nixb; \thetab)}$. 
Then, we condition the $i^{th}$ GP on the data that is sampled from the BCs/ICs imposed on the $i^{th}$ state variable. The $i^{th}$ conditional distribution is again a GP \cite{RN332,RN1559} and its expected value at the arbitrary query point $\nixb^*$ in the domain is:
%=================================================================
\begin{equation}
    \niz_i\parens{\nixb^*; \thetab, \phib_i} \coloneqq 
    \E\brackets{\niu_i^*|\niub_i, \nixb} = 
    m_i\parens{\nixb^*; \thetab} + 
    k_i\parens{\nixb^*, \niXb_i; \phib_i}k_i^{-1}\parens{\niXb_i, \niXb_i; \phib_i}
    \parens{\niub_i- m_i\parens{\niXb_i; \thetab}}
    \label{eq gp conditional mean}
\end{equation}
%=================================================================
where $i=1, ..., n_{\niu}$, $\niu_i^* = \niu_i(\nixb^*)$, $\niXb_i = \braces{\nixb^{(1)}, \cdots, \nixb^{(n)}}$ and the corresponding outputs $\niub_i = \braces{\niu_i(\nixb^{(1)}), \cdots, \niu_i(\nixb^{(n)})}$ denote the samples taken from the BCs/ICs for the $i^{th}$ state variable, $\thetab$ are the parameters of the mean function, $\boldsymbol{0} \leq \phib$ are the so-called length-scale or roughness parameters of the kernel, and $\zetab = \brackets{\thetab, \phib}$. 

The $\niub_i- m_i\parens{\niXb_i; \thetab}$ vector in \Cref{eq gp conditional mean} measures the error or \textit{residuals} of the mean function in reproducing the boundary data. This term is scaled by $k_i\parens{\nixb^*, \niXb_i; \phib_i}k_i^{-1}\parens{\niXb_i, \niXb_i; \phib_i}$ which are \textit{kernel-induced weights} that ensure $\niz_i\parens{\nixb^*; \thetab, \phib_i}$ interpolates $\niub_i$ regardless of the functional forms and parameters of $m_i(\nixb; \thetab)$ and $k_i(\nixb, \nixb'; \phib_i)$. This interpolating property is particularly attractive because it enables us to dispense \Cref{eq penalty opt generic} from the constraints that correspond to BCs/ICs as long as $n$ is large enough, i.e., as long as we sufficiently sample from BCs/ICs. Hence, \Cref{eq gp conditional mean} is our proposed parameterization for representing the state variables. 

Parameterization of the design variable or density is very similar to \Cref{eq gp conditional mean} but with $(1)$ updated data that correspond to the density, and $(2)$ application of the projection function $g(\cdot)$ that encourages binary outputs. That is:
%=================================================================
\begin{subequations}
    \begin{align}
        &\widetilde\niz_{-1}\parens{\nixb^*; \thetab, \phib_{-1}} \coloneqq 
        \E\brackets{\widetilde\rho^*|\rhob, \nixb} \notag \\
        &\quad= m_{-1}\parens{\nixb^*; \thetab} +  k_{-1}\parens{\nixb^*, \niXb_{-1}; \phib_{-1}}
        k_{-1}^{-1}\parens{\niXb_{-1}, \niXb_{-1}; \phib_{-1}}
        \parens{\rhob- m_{-1}\parens{\niXb_{-1}; \thetab}} \\
        &\niz_{-1}\parens{\nixb^*; \thetab, \phib_{-1}}  = g\parens{\widetilde\niz_{-1}\parens{\nixb^*; \thetab, \phib_{-1}}},
    \end{align}
    \label{eq gp conditional mean rho}
\end{subequations}
%=================================================================
where $\niXb_{-1} = \braces{\nixb^{(1)}, \cdots, \nixb^{(n)}}$\footnote{To simplify the descriptions and notation, we assume the density and state variables are known at $n$ locations but in our approach this number can vary across different variables.} and the corresponding density vector $\rhob = \braces{\rho(\nixb^{(1)}), \cdots,\rho(\nixb^{(n)})}$ denote the locations with known densities. \Cref{eq gp conditional mean rho} uses the shorthand notation where the subscript ${n_{\niu}+1}$ is replaced by the subscript $-1$.

While there are many kernels available \cite{yousefpour2024gp+}, we choose the following one in this work which is a simplified version of the Gaussian covariance function:
%=================================================================
\begin{equation} 
    k\parens{\nixb, \nixb'; \phib} = 
    \exp\braces{-\parens{\nixb - \nixb'}^T\diag(\phib)\parens{\nixb - \nixb'}} 
    + \mathbbm{1}\{\nixb==\nixb'\}\delta,
    \label{eq kernel}
\end{equation}
%=================================================================
where $\mathbbm{1}\{\cdot\}$ returns $1/0$ if the enclosed statement is true/false and $\delta$ is the so-called nugget parameter that is typically used to model noise or improve numerical stability of the covariance matrix $k\parens{\niXb, \niXb; \phib}$.

Based on \Cref{eq kernel}, the correlation between $\nixb$ and $\nixb'$ exponentially dies out as the weighted distance between them increases. Hence, as the distance between $\nixb^*$ and the boundary data $\niXb_i$ increases, the contributions of the second term on the right-hand side of \Cref{eq gp conditional mean} to $\niz_i\parens{\nixb^*; \thetab, \phib_i}$ decreases. The pace of this decrease along the $k^{th}$ direction in the design domain directly depends on the $k^{th}$ elements in $\phib_i$. In this work, we require $\niz_i\parens{\nixb^*; \thetab, \phib_i}$ to interpolate the boundary data which requires $\phib_i$ to be sufficiently large. With this choice, $\niz_i\parens{\nixb^*; \thetab, \phib_i}$ primarily depends on $m_i\parens{\nixb^*; \thetab}$ if $\nixb^*$ is not on $\partial\Omega$ because the kernel-induced weights become zero. 

\begin{figure*}[!hbt]
    \centering
    \includegraphics[width=1.00\columnwidth]{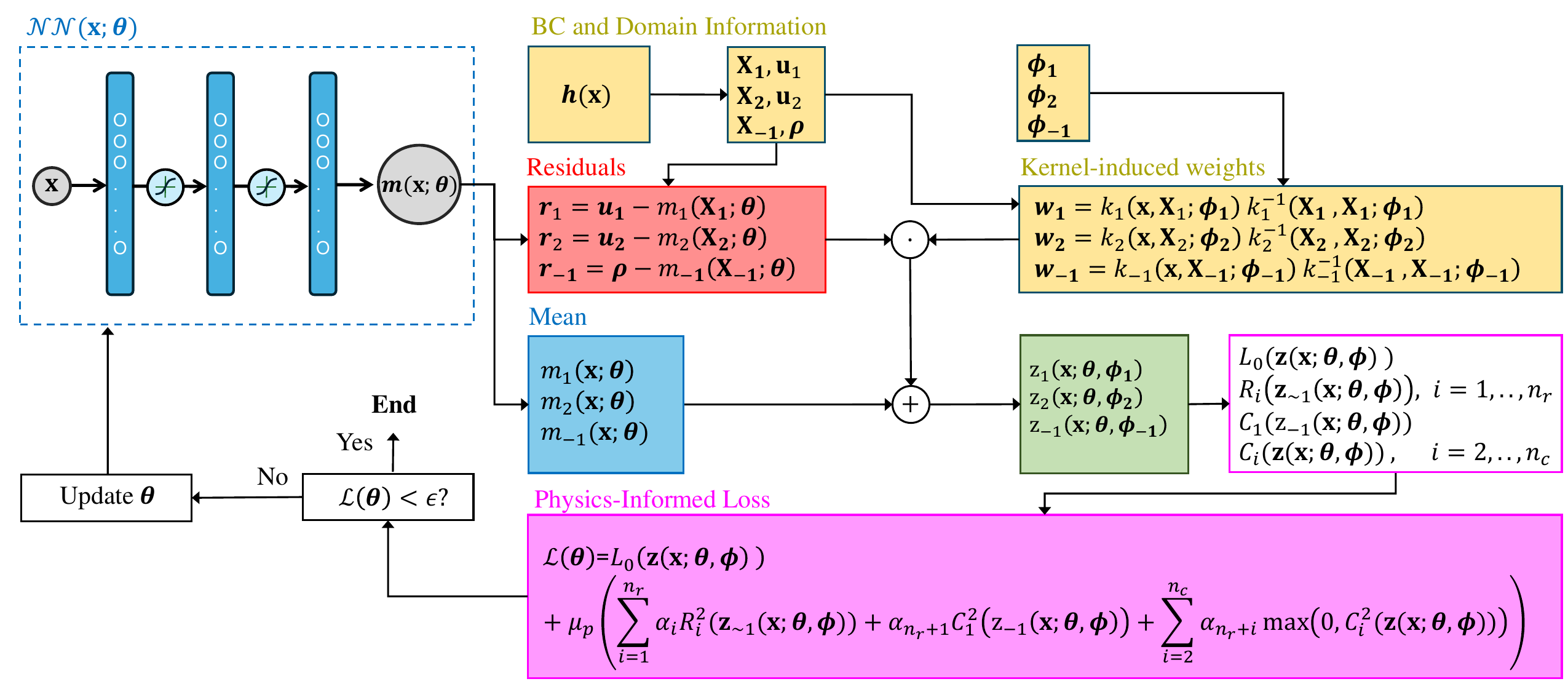}
    \caption{\textbf{Simultaneous and meshfree topology optimization:} $\rho(\nixb)$ denotes the design variable and it is assumed that the structure has two state variables $\niub(\nixb) = \brackets{u_1(\nixb), u_2(\nixb)}$. The covariance matrices ensure that the variables satisfy the boundary conditions while the parameters of the mean function are optimized to minimize \Cref{eq penalty opt generic}. In practice, we fix the length-scale parameters of all the kernels to $10^2$ and only optimize $\thetab$ via \Cref{eq penalty opt generic 2}.}
    \label{fig flowchart}
\end{figure*}

The flowchart of our approach is schematically illustrated in \Cref{fig flowchart} and demonstrates that the mean function of the GPs is responsible for approximating $\niub(\nixb)$ and $\rho(\nixb)$ inside the domain while their independent kernel-based components ensure that the state variables and the density function satisfy the imposed constraints. As detailed in \Cref{subsec training}, $\phib$ and $\thetab$ can be efficiently estimated by, respectively, simple heuristics and minimizing a slightly modified version of \Cref{eq penalty opt generic}.

%=================================================================
%=================================================================

As shown in \Cref{fig flowchart} and noted before, $\nizb\parens{\nixb^*; \thetab, \phib}$ has a single multi-output mean function but as many kernels as there are state and design variables, that is $\niub(\nixb)$ and $\rho(\nixb)$. The rationale behind the latter decision is that it dramatically reduces the computational costs and memory footprint of the approach since the covariance matrices in \Cref{eq gp conditional mean} are only built for one dependent variable. For instance, for a problem with $m$ dependent variables where each one is sampled at $n$ boundary locations, our choice requires working with $m$ covariance matrices of size $n\times n$ instead of working with a single covariance matrix of size $mn\times mn$.

After plugging $\nizb\parens{\nixb^*; \thetab, \phib}$ into \Cref{eq penalty opt generic} and minimizing $\losszeta$, the shared mean function automatically encodes the relations between the state and design variables. Since the constraints in \Cref{eq penalty opt generic} are typically nonlinear functions of these variables, the conditional distributions $\nizb\parens{\nixb^*; \widehat\thetab, \widehat\phib}$ are no longer GPs and merely a maximum a posteriori estimate \cite{RN1886}. This non-Gaussianity does not cause any issues in our TO approach because we only rely on the fact that the formulation in \Cref{eq gp conditional mean} can interpolate the boundary data.

We conclude this section by highlighting some connections between our method with the level set approach where the density is obtained by projecting the level set function to a binary variable. Referring to \Cref{fig flowchart} we observe that the intermediate density function (the output of the deep NN) is obtained by linearly projecting the last hidden layer of the NN, which is high dimensional, into a scalar variable which is then plugged in \Cref{eq gp conditional mean} and finally passed through the projection function $g(\cdot)$. This process of obtaining the density based on a high-dimensional vector is quite similar in nature to how the density is obtained in two-field SIMP or level set even though our TO approach is fundamentally different from these two methods.

\subsection{Efficient Parameter Estimation} \label{subsec training}

The parameters of our model can be collectively estimated via \Cref{eq penalty opt generic} with its individual components defined in \Cref{eq penalty opt details}. In this section, we introduce three mechanisms to dramatically accelerate and stabilize this optimization process. First, we note that the parameterization in \Cref{eq gp conditional mean} involves the two covariance matrices $k_i\parens{\nixb^*, \niXb_i; \phib_i}$ and $k_i^{-1}\parens{\niXb_i, \niXb_i; \phib_i}$ where $i$ indexes state and design variables. The optimization process in \Cref{eq penalty opt generic} requires repeated construction of these two matrices (one of which involves matrix inversion) and their multiplications with other vectors in \Cref{eq gp conditional mean}. Our choice for endowing each of these variables with an independent kernel reduces the size of these covariance matrices and thus decreases the evaluation cost of $\losszeta$. However, the overall cost of minimizing \Cref{eq penalty opt generic} can still be substantially high because it relies on repeatedly evaluating $\losszeta$ and its gradients with respect to $\thetab, \phib,$ and $\nixb$.

In addition to high costs, a potential issue with \Cref{eq gp conditional mean} is that $k_i\parens{\niXb_i, \niXb_i; \phib_i}$ can become ill-conditioned as $\phib$ is updated during the optimization. While the nugget parameter can alleviate this numerical instability, it does so at the expense of increasing the costs\footnote{Finding the appropriate nugget parameter is itself an iterative process.} and preventing interpolation which is essential to our approach. 

To address the above issues on estimating $\phib$, we propose to fix it to a relatively large value\footnote{In all of our studies, we use $10^2$.} to ensure $k_i\parens{\niXb_i, \niXb_i; \phib_i}$ is numerically stable and \Cref{eq gp conditional mean,eq gp conditional mean rho} can strictly interpolate $\niub_i$ and $\rhob$, respectively. Hence, as shown in \Cref{fig flowchart} the optimization problem reduces to estimating the parameters of the deep NN, that is:
% ============================================================= %
\begin{equation}
    \begin{aligned}
        \widehat\thetab = \underset{\thetab}{\argmin} \hspace{2mm} \losstheta
        &= \underset{\thetab}{\argmin} \hspace{2mm} 
        \Obj\parens{\nizb(\nixb; \thetab, \widehat\phib)} + 
        \mu_p \left( 
        \sum_{i=1}^{n_r}\alpha_i R_i^2\parens{\nizb_{\sim1}(\nixb; \thetab, \widehat\phib)} \right. \\
        & \left. + \alpha_{n_r + 1}C_1^2\parens{\niz_{-1}(\nixb; \thetab, \widehat\phib)} + 
        \sum_{i=2}^{n_c}\alpha_{n_r+i} \max\parens{0, C_i^2\parens{\nizb(\nixb; \thetab, \widehat\phib)}} \right)
    \end{aligned}
    \label{eq penalty opt generic 2}
\end{equation}
% ============================================================= %
where $\mu_p$ represents the penalty coefficient which is incrementally increased during the optimization (up to a maximum value) to ensure the final design satisfies the constraints \cite{bertsekas2014constrained}. In our studies, we use the following simple mechanism to increase $\mu_p$ every $50$ epochs until it reaches a maximum value of $500$:
% ============================================================= %
\begin{equation}
    \mu_p^{k+1} = \beta_p \mu_p^k,
    \label{eq penalty opt 2}
\end{equation}
% ============================================================= %
where $\beta_p = 1.05$ is the growth factor and $\mu_p^0 = 1$.

The choice of fixing $\phib$ is consistent with our rationale for parameterizing the state and design variables with \Cref{eq gp conditional mean,eq gp conditional mean rho} which relies on the kernels to merely interpolate the sampled boundary data.
Additionally, fixing $\phib$ allows us to build, invert, and store all the covariance matrices so that they can be repeatedly used while minimizing $\losstheta$.

The second and third mechanisms that we develop to accelerate the optimization problem in \Cref{eq penalty opt generic 2} are based on gradient calculations and balancing the individual components of $\losstheta$. We explain these two mechanisms in \Cref{subsubsec finite difference,subsubsec loss weights}. 

\subsubsection{High-Order Finite Difference} \label{subsubsec finite difference}

The gradients of $\losstheta$ with respect to $\thetab$ can be analytically calculated via the chain rule and automatic differentiation since all the components of $\nizb\parens{\nixb^*; \thetab, \phib}$ and the operations in \Cref{eq penalty opt generic 2} are differentiable. Hence, we can obtain $\widehat\thetab$ by using a gradient-based optimizer such as Adam which is a first order method. 
Specifically, we first initialize the deep NN parameters as $\thetab_k$ to obtain the state and design variables at $n$ CPs distributed in $\Omega$, i.e., $\nizb\parens{\niXb^*; \thetab_k, \widehat\phib}$ where $\niXb^*$ denotes the CPs. Then, we calculate the average gradients of $\losstheta$ with respect to $\thetab$ at $\niXb^*$, i.e., $\frac{1}{n}\sum_{j=1}^n \nabla_{\thetab} \losstheta$, and use them to update the initial set of parameters. These updates are repeated for a predefined number of iterations at which point $\widehat\thetab$ is obtained.

The major computational bottleneck in calculating $\nabla_{\thetab} \losstheta$ are the constraints that correspond to the state equations, i.e., $R_i^2\parens{\nizb_{\sim1}(\nixb; \thetab, \widehat\phib)}$ in \Cref{eq penalty opt generic 2}. These constraints are based on differential equations that involve high-order gradients of $\nizb_{\sim1}(\nixb; \thetab, \widehat\phib)$ with respect to $\nixb$. Referring to the computational graph in \Cref{fig flowchart} we observe that the operations that connect $\nixb$ to $R_i^2\parens{\nizb_{\sim1}(\nixb; \thetab, \widehat\phib)}$ are very deep and complex. Such connections make it very expensive to use the chain rule to calculate both $\nabla_{\thetab} \losstheta$ and $R_i^2\parens{\nizb_{\sim1}(\nixb; \thetab, \widehat\phib)}$ for each optimization iteration.
We highlight that depending on the application this computational issue may involve other terms on the right-hand side of \Cref{eq penalty opt generic 2}. For instance, estimating the performance metric $\Obj\parens{\nizb(\nixb; \thetab, \widehat\phib)}$ in \Cref{sec results} also involves partial derivatives with respect to $\nixb$. 

To address the computational bottlenecks associated with gradient calculations, we position $\niXb^*$ on a regular grid and use high-order finite difference (FD) to estimate the gradients with respect to $\nixb$.
Specifically, at the $i^{th}$ step of the optimization we use $(1)$ FD to calculate all the terms on the right-hand side of \Cref{eq penalty opt generic 2} whose evaluation relies on differentiation with respect to $\nixb$, e.g., $\nabla_{\nixb} \nizb_{\sim1}(\niXb^*; \thetab_i, \widehat\phib)$, and $(2)$ chain rule to obtain $\nabla_{\thetab} \losstheta$ and update $\thetab_k$ where $k$ denotes the iteration number.
This mixed approach for differentiation accelerates the parameter estimation process while negligibly adding errors into the calculations. We argue that these errors minimally affect the results since their scales are much smaller than the residuals that control the continuation nature of our TO framework. The results in \Cref{sec results} support this argument.

To showcase the above process, we provide the formulations for FE-based gradient calculations for the application studied in \Cref{sec results} for which $\nixb = \brackets{x, y}$ and $\niub(\nixb) = \brackets{u(x, y), v(x, y), p(x, y), \rho(x, y)}$. $\brackets{x, y}$ are the spatial coordinates of the domain which we discretize into a regular grid via $dx\times dy$ elements, $u(x, y)$ and $v(x, y)$ denote the horizontal and vertical velocity fields, and $p(x, y)$ is the pressure.
Taking $u(x, y)$ as an example, we approximate the first-order spatial derivatives via the following equations which have $4^{th}$ order accuracy:
%=================================================================
\begin{equation}
    \begin{aligned}
    \frac{\partial u}{\partial x} &\approx \frac{-u_{i+2,j} + 8u_{i+1,j} - 8u_{i-1,j} + u_{i-2,j}}{12dx}, \\
    \frac{\partial u}{\partial y} &\approx \frac{-u_{i,j+2} + 8u_{i,j+1} - 8u_{i,j-1} + u_{i,j-2}}{12dy}.
    \end{aligned}
    \label{eq_first_order_derivatives_u_higher_order}
\end{equation}
%=================================================================
where $i$ and $j$ index the points in the grid. 
The Laplacian of \( u(x, y) \) which involves its second-order derivatives is calculated as:
%=================================================================
\begin{equation}
    \begin{aligned}
        \nabla^2 u \approx &\frac{-u_{i+2,j} + 16u_{i+1,j} - 30u_{i,j} + 16u_{i-1,j} - u_{i-2,j}}{12dx^2} + \\
        & \frac{-u_{i,j+2} + 16u_{i,j+1} - 30u_{i,j} + 16u_{i,j-1} - u_{i,j-2}}{12dy^2}.        
    \end{aligned}
    \label{eq_second_order_derivatives_u_higher_order}
\end{equation}
%=================================================================

To handle boundary conditions effectively, we employ ghost points. These ghost points are positioned outside the boundary and assigned values according to the BCs.

\subsubsection{Dynamic Loss Weighting} \label{subsubsec loss weights}

The individual loss components that serve as penalties on the right-hand side of \Cref{eq penalty opt generic 2} typically have different scales. Additionally, the magnitude of some of the penalty terms can change quite dramatically as $\thetab$ is updated during the optimization. This is especially the case for the constraints that correspond to state equations involving high-order gradients which can change substantially for small changes in $\thetab$. 

To ensure the individual penalty terms in \Cref{eq penalty opt generic 2} have similar scales during the entire optimization process, we develop an adaptive mechanism to adjust their weights. Motivated by \cite{shishehbor2024parametric, wang2021understanding}, we first choose one of the loss terms as a reference and assign the unit weight to it. Then, we determine the weights of the other terms such that their gradients with respect to $\thetab$ have comparable magnitudes at any iteration during the optimization; preventing any loss term from dominating $\nabla_{\thetab}\losstheta$ which is used for updating $\thetab$.

To achieve this, we use an Adam-inspired weight balancing approach. Specifically, at each iteration of the optimization, we update the weight for each loss term as follows:
%=================================================================
\begin{equation}
    \alpha_{i}^{k+1} = (1 - \lambda) \alpha_i^{k} + \lambda \frac{\max \left| \nabla_\theta \mathcal{L}_r(\theta) \right|}{\mathrm{mean} \left| \nabla_\theta \mathcal{L}_i(\theta) \right|},
    \label{eq adam inspired}
\end{equation}
%=================================================================
where $\alpha_i^{k+1}$ denotes the weight of the loss term $\mathcal{L}_i$ in the $(k+1)$-th iteration of the optimization process. $\nabla_\theta \mathcal{L}_i(\theta)$ denotes the gradient of the loss term $\mathcal{L}_i$ with respect to $\theta$ and the term $\mathcal{L}_r$ represents the reference loss term. The parameter $\lambda$ is typically set to $0.9$. By adding a small value $\epsilon$ to the denominator of \Cref{eq adam inspired}, we bias-correct the weights away from zero and prevent numerical instabilities.

\Cref{alg: TopOp-algorithm} summarizes our framework for simultaneous and meshfree TO which has a single for loop and does not discretize the design domain and variables.
%=================================================
\begin{algorithm} [h!]
    \SetAlgoLined
    \DontPrintSemicolon
    \textbf{Given:} Design domain, imposed boundary conditions, objective function $\Obj\parens{\nizb(\nixb; \zetab)}$, constraints \\  
    \textbf{Define:}
\begin{itemize} [label={}]
    \setlength\itemsep{0em}
    \item - Multi-output function $\nizb(\nixb; \zetab)$, \Cref{eq z definition}
    \item - $C_1\parens{ \niz_{-1}(\nixb; \zetab)}$, $C_i\parens{\nizb(\nixb; \zetab)}$, $R_i\parens{\nizb_{\sim1}(\nixb; \zetab)}$,  \Cref{eq penalty opt details}
    \item - GP priors on $\nizb(\nixb; \zetab)$ with independent kernels $k(\nixb, \nixb'; \phib)$ and shared mean function $\mb(\nixb; \thetab)$
    \item - Physics informed loss functions $\losszeta$, \Cref{eq penalty opt generic}
    \item - Stop condition
\end{itemize}
    \textbf{Specify:} Kernel parameters $\phib$, static loss weights, penalty coefficient\\
    \While{stop conditions not met}{
    \begin{enumerate}
        \item Calculate $\nizb\parens{\niXb^*; \thetab_i, \widehat\phib}$, \Cref{eq gp conditional mean,eq gp conditional mean rho}  
        \item Calculate loss function  $\lossthetak{k}$, \Cref{eq penalty opt generic 2} 
        \item Update trainable parameters $\thetab^{k+1} \leftarrow \thetab^{k}$
        \item Update loss weights: $\alphab^{k+1} \leftarrow \alphab^{k}$, \Cref{eq adam inspired}
        \item Update penalty coefficient weights: $\mu_p^{k+1} \leftarrow \mu_p^{k}$, \Cref{eq penalty opt 2}
    \end{enumerate}       
    }
    \textbf{Output:} Multi-output function $\nizb(\nixb; \zetab)$ 
 
    \caption{Simultaneous and meshfree topology optimization with physics-informed GPs}
    \label{alg: TopOp-algorithm}
\end{algorithm}
%=================================================

    \section{Results and Discussions} \label{sec results}

\begin{figure*}[!b]
    \centering
    \begin{subfigure}[b]{0.48\textwidth}
        \centering
        \includegraphics[width=1.00\columnwidth]{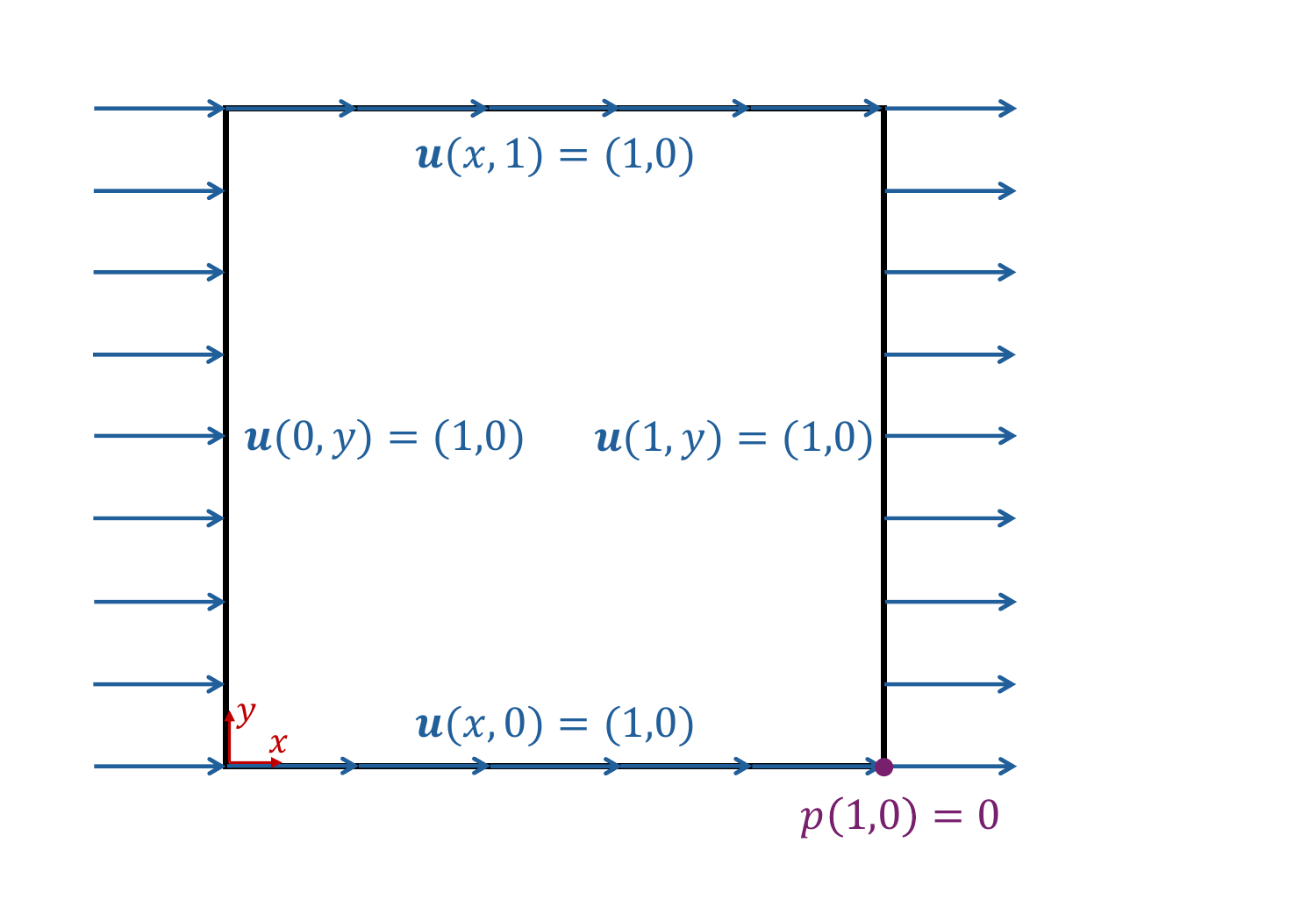}
        \caption{Rugby}
        \label{fig:sub1}
    \end{subfigure}
    \hfill
    \begin{subfigure}[b]{0.48\textwidth}
        \centering
        \includegraphics[width=1.00\columnwidth]{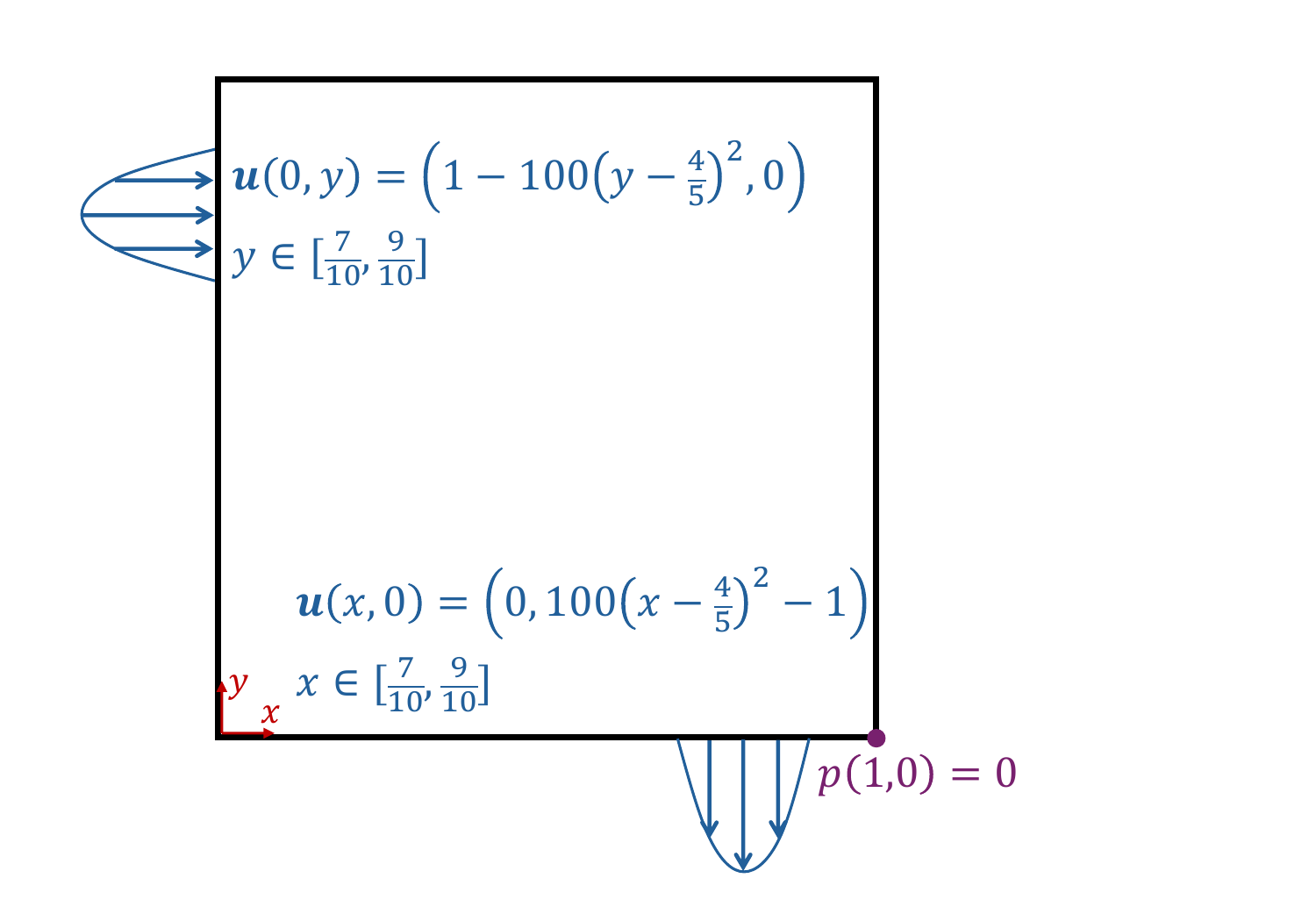}
        \caption{Pipe bend}
        \label{fig:sub2}
    \end{subfigure}
    \vfill
    \begin{subfigure}[b]{0.48\textwidth}
        \centering
        \includegraphics[width=1.00\columnwidth]{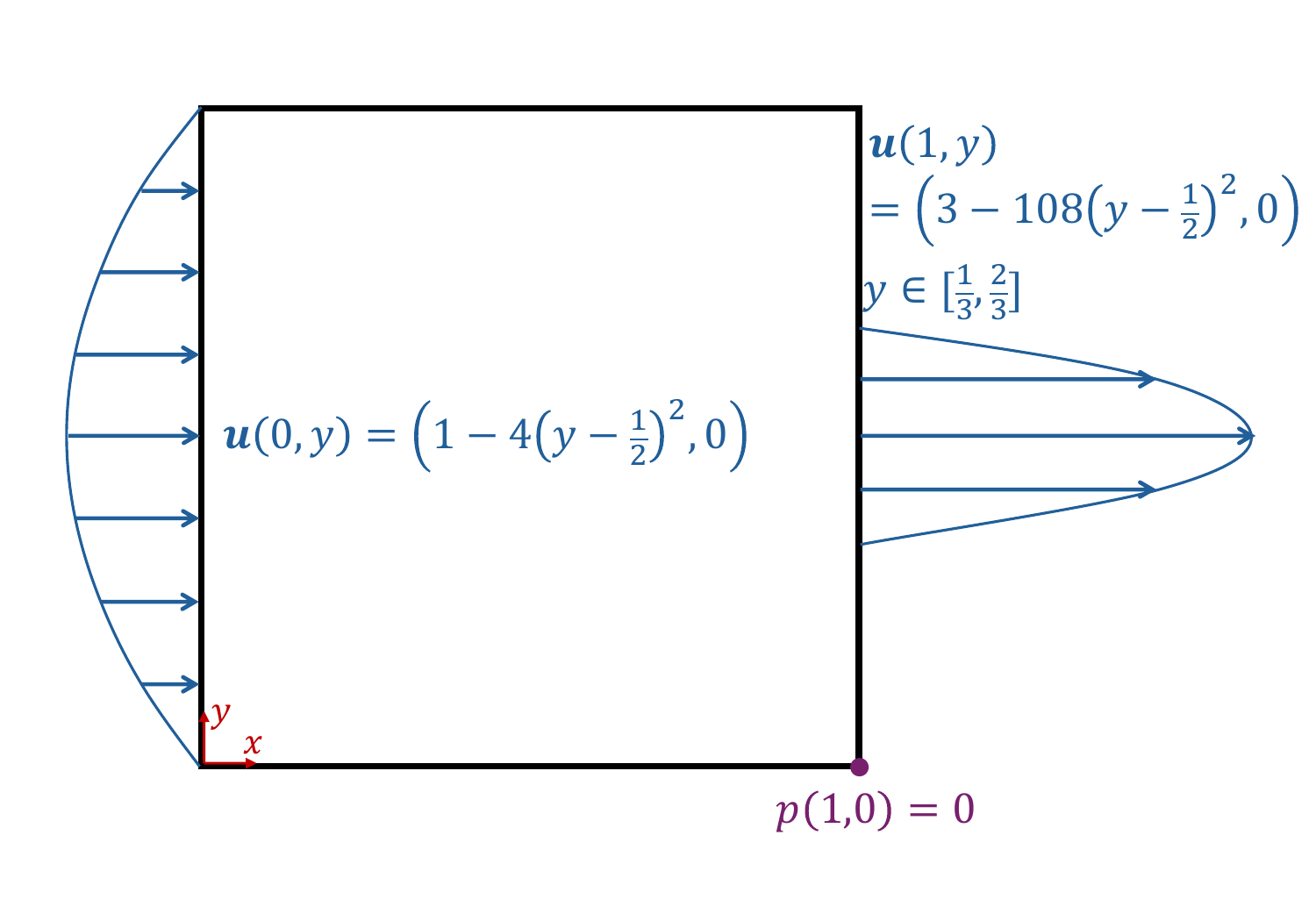}
        \caption{Diffuser}
        \label{fig:sub3}
    \end{subfigure}
    \hfill
    \begin{subfigure}[b]{0.48\textwidth}
        \centering
        \includegraphics[width=1.00\columnwidth]{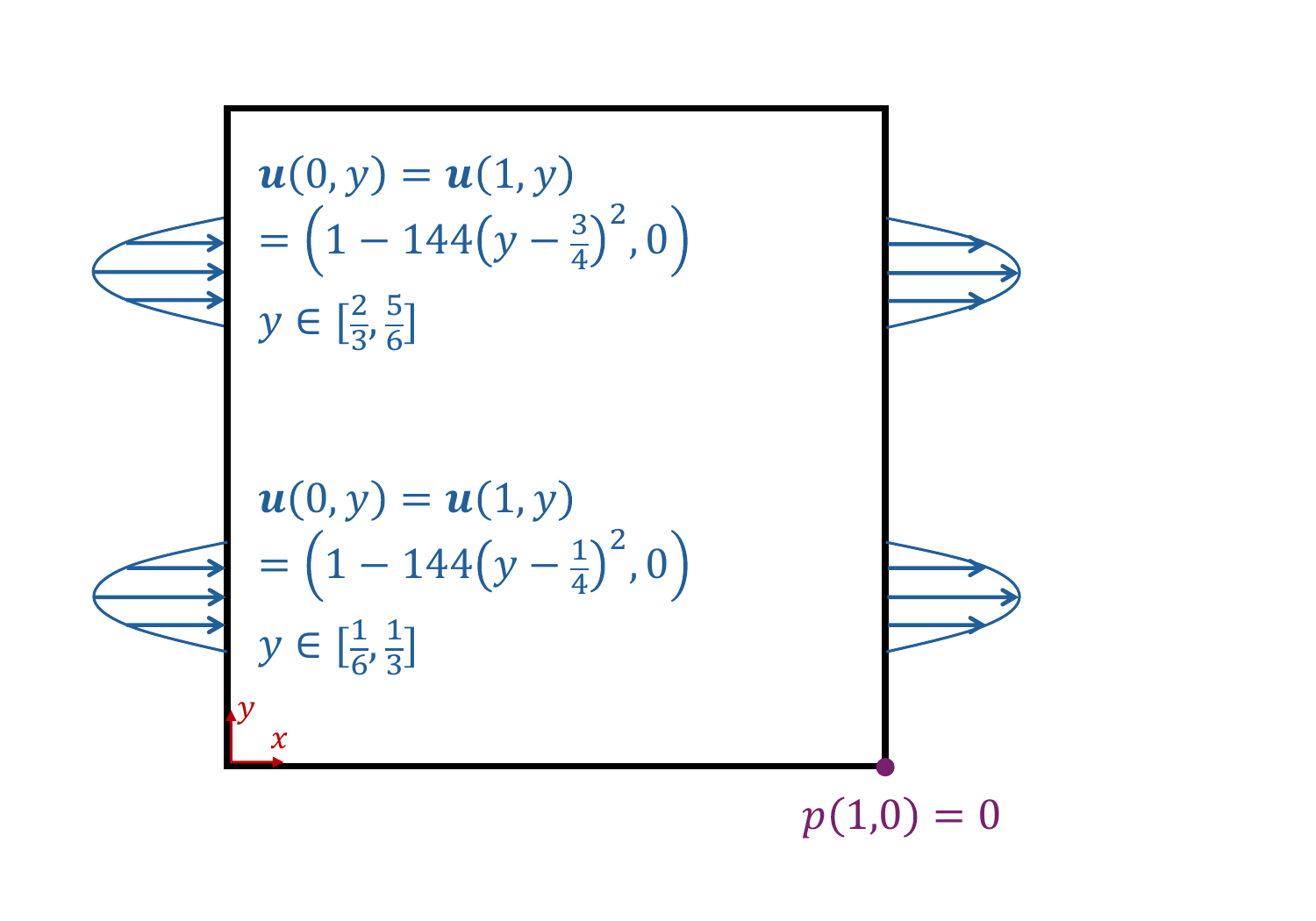}
        \caption{Double pipe}
        \label{fig:sub4}
    \end{subfigure}
    \caption{\textbf{Design domain and the imposed boundary conditions:} The domain is a unit square in all cases and the boundary conditions are shown in each case. Pressure is known at only one point in all cases.}
     \label{fig four problems}
\end{figure*}

% Flow should be like this:
% Describe the problem (you can use the original paper by Borvall for this), include equations and explain what each equation represents. 
We study the four problems introduced in \cite{RN1986} where the goal is to identify the spatial material distribution that minimizes the flow's dissipated power subject to a predefined volume constraint.
The problem is defined under the incompressible flow assumption for Newtonian fluids for which the momentum equation reads as:
%=========================================================
\begin{equation}
    \varrho\left(\boldsymbol{u}_t+\boldsymbol{u} \cdot \nabla \boldsymbol{u}\right)=-\nabla p+\mu \nabla^2 \boldsymbol{u}+\boldsymbol{f} , 
    \label{eq: NS-momentum}
\end{equation}
%==========================================================
where $\boldsymbol{u}$ is the velocity field, $\nabla{\boldsymbol{u}}$ is the velocity gradient, $p$ is pressure, $\varrho$ and $\mu$ are the fluid's density and dynamic viscosity, respectively, and $\boldsymbol{f}$ denotes the body force per unit volume. 
Based on Darcy's law, the porous medium exerts a drag force proportional to the viscous fluid's velocity, which could be treated as a body force. Assuming negligible gravitational forces, $\boldsymbol{f}$ in \Cref{eq: NS-momentum} only accounts for the resistive effects of the porous medium with the inverted permeability $\kappa^{-1}$. That is:
%=========================================================
\begin{equation}
    \boldsymbol{f} = - \mu \kappa^{-1} \boldsymbol{u}.
    \label{eq: darcy-frag}
\end{equation}
%==========================================================
Additionally, for flow in porous media at low Reynolds numbers, the inertial term on the left-hand side in \Cref{eq: NS-momentum} can be neglected. Thus, the Navier-Stokes equations are simplified to the steady-state Brinkman equations \cite{brinkman1949calculation}:
%=================================================================
\begin{subequations}
    \begin{align}
        &  -\nabla p+\mu \nabla^2 \boldsymbol{u} - \mu \kappa^{-1} \boldsymbol{u} = 0
        \label{eq: momentum}\\
        & \nabla \cdot \boldsymbol{u} = 0
        \label{eq: continuity}.
    \end{align}
    \label{eq: Brinkman}
\end{subequations}
%=================================================================
Power dissipation in the porous medium is caused by the work done by the viscous and drag forces and hence our objective function considering $\mu = 1$ in a 2D domain is defined as follows:
%=================================================================
\begin{equation}
    \mathcal{J}= \frac{1}{2} \int_{\Omega}\left( \nabla \boldsymbol{u}: \nabla \boldsymbol{u} +\kappa^{-1} \left \| \boldsymbol{u} \right \| ^2\right) dx dy ,
    \label{eq: objective}
\end{equation}
%=================================================================
where the design variable $\rho$ is related to the inverted permeability by \cite{RN1986}:
%==================================================

\begin{equation}
    \kappa^{-1}(\rho)= {\kappa^{-1}}_{max} + \left({\kappa^{-1}}_{min} - {\kappa^{-1}}_{max}\right) \rho  \frac{1 + q}{\rho+q},
    \label{eq: alpha_ours}
\end{equation}
%=============================================
where ${\kappa^{-1}}_{max} = 2.5 \times 10^4 , \kappa^{-1}_{min} =  2.5 \times 10^{-4}$, and $q = 0.1$. The optimization problem that minimizes $\mathcal{J}$ is subject to the equality constraint prescribing the fluid volume fraction: 
%=================================================================
\begin{equation}
    \int_{\Omega}\rho dx dy = V , 
    \label{eq: constraint}
\end{equation}
%=================================================================
where $V$ is equal to $0.9 , 0.08\pi, 0.5$ and $\sfrac{1}{3}$ for our four test cases which are named as Rugby, Pipe bend, Diffuser, and Double pipe, respectively. We demonstrate the BCs corresponding to each of these cases in \Cref{fig four problems} where domain boundaries (excluding inlets/outlets) represent no-slip walls.

%=====================================================================

%===================================================

We use the following settings (which are unchanged across the four examples) to minimize \Cref{eq: objective} for each of the cases in \Cref{fig four problems}.
To ensure the BCs are accurately captured while the computational costs of matrix multiplications in \Cref{eq gp conditional mean,eq gp conditional mean rho} are minimized, we sample $n_{bc} = 25$ points from the BCs on each side of the domain.
For calculating the PDE residuals, we sample the domain with $n_{cp} = 100^2$ CPs which are structured on a regular grid to facilitate the FD-based accelerations described in \Cref{subsubsec finite difference}. We use $\alpha_1 = \alpha_2 = 1$ since $R_1$ and $R_2$ have a similar nature and set the projection function to $g(\widetilde\niz_{-1}) = (1+\exp(-12(\widetilde\niz_{-1} - 0.5)))^{-1}$. We use Adam optimizer to train our models with an initial learning rate of $0.001$ which is scheduled to decay by $0.75$ four times during $50k$ epochs.
% In addition to the boundary conditions shown in \Cref{fig four problems}, in the rugby case, our model treats the center of the domain $\nixb_{center} = [0.5 , 0.5]^\top$ as a solid stagnation point. (i.e., $\rho (\nixb_{center}) = 0 , \boldsymbol{u}(\nixb_{center}) = (0,0)$). 

In COMSOL, we set up the four problems via the Brinkman Equations (br) interface and employ the density method. We introduce a control variable field $\bar{\rho}(\nixb) \in [0,1]$ to represent the material distribution in the domain and interpolate it via SIMP as:
%==================================================
\begin{equation}
    \rho(\nixb) = \rho_{min} + (1 - \rho_{min})\bar{\rho}^\gamma(\nixb),
    \label{eq: simp}
\end{equation}
%=============================================
where $\rho_{min} = 0.001$ and $\gamma = 2$ are the minimum penalized volume fraction and SIMP exponent, respectively. 
In contrast to compliance minimization problems where a SIMP exponent of $3$ is typically adopted, we obtained the best results from COMSOL via $\gamma = 2$.
Additionally, in the Pipe bend and Double pipe examples, a Helmholtz-type filter as in \Cref{eq: helmholtz} with a filter radius $r = 0.01$ is applied to $\bar{\rho}$ before interpolation to prevent optimization failure or numerical instabilities. This filter is not used in Rugby and Diffuser problems as it reduces the performance (in terms of minimizing the dissipated power) of the optimized topology. 

\Cref{eq: alpha_ours} does not robustly work in COMSOL so we use the following relation which is proposed in \cite{DUAN20161131}:
%==================================================
\begin{equation}
    \kappa^{-1}(\rho)= {\kappa^{-1}}_{min} + \left({\kappa^{-1}}_{max} - {\kappa^{-1}}_{min}\right) q  \frac{1 - \rho}{q + \rho},
    \label{eq: alpha_comsol}
\end{equation}
%=============================================
where ${\kappa^{-1}}_{min} = 0$ , ${\kappa^{-1}}_{max} = 10^4$ and $q = 0.2$ are chosen empirically. With our approach, both relations in \Cref{eq: alpha_ours,eq: alpha_comsol} work but we choose the former to be consistent with the original reference \cite{RN1986}.

We choose the Interior Point Optimizer (IPOPT) coupled with Parallel Direct Sparse Solver Interface (PARDISO) implemented in COMSOL to solve the optimization problem with $0.001$ optimality tolerance and $200$ maximum iterations, with the exception of Diffuser, for which the maximum iteration number is set to $50$. 
To be consistent with our model's discretization, we use $100 \times 100$ regular grid mesh for Rugby and a slightly distorted grid of quadrilateral elements with roughly the same total number of elements for other cases to allow for accurate positioning of boundary nodes.

Unlike our approach which always starts with a random initial topology (since the parameters of the mean function are initialized randomly), the initial topology can be directly designed in COMSOL. So, we consider two initial topologies for COMSOL in our comparative studies to see the effects of initialization on the results. Specifically, we consider both random and constant initialization where in the first case the density of each element is drawn from a standard uniform distribution while in the latter case the density values are all set to $0.5$.
We repeat the analysis $10$ times for both our approach and COMSOL with random initialization. 

% IPOPT utilizes an interior point line search filter method to find a local solution to nonlinear programming problems (NLPs). It leverages first and second derivative information and uses quasi-Newton methods to approximate Hessians. Further algorithmic details could be found in \cite{nocedal2009adaptive,wachter2003line,wachter2005line,wachter2006implementation,wachter2002interior,schenk2001pardiso}.  

\subsection{Summary of Comparative Studies} \label{subsec summary results}
The results of our comparative studies are summarized in \Cref{tab: results-summary} with more details included in \Cref{tab: stats} in the case of random initializations. We observe in \Cref{tab: results-summary} that our approach provides similar objective values to both versions of COMSOL except in the case of Rugby example where we clearly outperform. Specifically, in this example COMSOL converges to local minima when initialized with either the constant value or randomly for $6$ out of $10$ seed numbers. However, our approach achieves the globally optimal topology in all $10$ training repetitions and hence provides a significantly lower median value for the objective function. 

%=================================================================
\begin{table*}[!b]
    \centering
    \renewcommand{\arraystretch}{1.5} % Adjust the cell height multiplier as needed
    %\footnotesize
    
    \setlength\tabcolsep{3pt} % Adjust cell separation as needed to fit new columns
    \begin{tabular}{l|cc|c|cc|cc}
    \hline
    \textbf{Problem} & \multicolumn{3}{c|}{\textbf{Our Approach}} & \multicolumn{2}{c|}{\textbf{COMSOL} - Random} & \multicolumn{2}{c}{\textbf{COMSOL} - Constant} \\ 
    & $\mathcal{J}$ & Time (sec) & $\mathcal{J}_c$ & $\mathcal{J}$ & Time (sec) & $\mathcal{J}$ & Time (sec) \\ \hline
    \textbf{Rugby} & 13.970 & 1288 & 13.924 & 20.326 & 741 & 22.300 & 359 \\ \hline
    \textbf{Pipe bend} & 9.972 & 1309 & 10.213 & 10.192 & 260 & 9.902 & 349 \\ \hline
    \textbf{Diffuser} & 30.749 & 1289 & 30.071 & 30.161 & 358 & 29.482 & 2504 \\ \hline
    \textbf{Double pipe} & 22.944 & 1292 & 23.329 & 21.475 & 568 & 20.797 & 870 \\ \hline
    \end{tabular}
    \caption{\textbf{Summary of the results} The median of final objective function value and computational cost across $10$ repetitions are reported for our model and COMSOL with random initialization. The performance of COMSOL with constant initialization is independent of seed and therefore the reported values are for one optimization instance. The objective function value $\mathcal{J}_c$ is computed by COMSOL for our approach's final median optimal topology. See the discussions in \Cref{subsec residuals} regarding $\mathcal{J}_c$.}
    \label{tab: results-summary}
\end{table*}
%=================================================================

We report the optimization time associated with all cases in \Cref{tab: results-summary} while acknowledging that the computational costs of COMSOL and our approach are not directly comparable due to inherent differences in hardware utilization and the programming languages upon which they are built (our approach is implemented in Python while COMSOL is based on C/C++ and Java). Our model's computational cost remains consistent across all cases, ranging from $1289$ to $1309$ seconds on NVIDIA RTX 4090 GPU. COMSOL with random initialization on AMD Ryzen 7 6800HS CPU is $1.8$ to $5$ times faster. Interestingly, we observe that COMSOL's efficiency depends on the initial topology with the costs generally increasing when the density vector $\rhob$ is initialized at $0.5$, especially in the case of Diffuser problem where the cost is increased by about $7$ times. 

Looking at the detailed statistics in \Cref{tab: stats} we observe that the median values reported in \Cref{tab: results-summary} are representative, i.e., the results provided by both methods (i.e., our approach and COMSOL with random initialization) are quite insensitive to the randomness. While not reported for brevity, a similar trend is observed for the computational costs of both methods in that they insignificantly change across the random repetitions (note that in our approach the ``consistent" computational cost applies to all problems as well as different repetitions in a specific problem. However, in the case of COMSOL, the cost varies across different problems).

Considering the results in \Cref{tab: results-summary} one may argue that, except in the case of the Rugby problem, our approach does not provide a substantial advantage over COMSOL especially since the latter is computationally faster (bearing in mind the differences in hardware and software utilization between the two approaches). In response to this argument, we highlight the following features of our approach. First, COMSOL requires some fine-tuning across the four problems because $(1)$ the Helmholtz filter in \Cref{eq: helmholtz} is used in the Pipe bend and Double pipe problems to improve numerical stability, and $(2)$ convergence of the optimization depends on how the density and permeability are related, i.e., COMSOL works with \Cref{eq: alpha_comsol} but not with \Cref{eq: alpha_ours} which was originally used in \cite{RN1986}. Unlike COMSOL, our approach does not require any fine-tuning across the four problems studied in this paper.

Second, we solve the constrained optimization problem via the penalty method and use Adam optimizer with a simple termination metric (i.e., $50k$ epochs). These choices simplify the implementation and are effective but as illustrated in \Cref{subsec topology evolution} they $(1)$ result in very slow convergence where the topology and the value of the objective function insignificantly change after about $20k$ epochs, and $(2)$ make our approach generally more conservative and more reliable than COMSOL in terms of satisfying the design constraint on volume fraction. 

%=================================================================
\begin{table*}[!t]
    \centering
    \begin{tabular}{lcccccccccc}
        \toprule
        \textbf{Problem} & mean & median & std & min & max & mean & median & std & min & max\\
        \midrule
        & \multicolumn{5}{c}{\textbf{Our Approach}} & \multicolumn{5}{c}{\textbf{COMSOL}}\\        

        \cmidrule(lr){2-6} 
        \cmidrule(lr){7-11} 
        
        \textbf{Rugby} & 14.05 & 13.97 & 0.43 & 13.58 & 15.03 & 20.61 & 20.33 & 5.72 & 14.17 & 28.10\\
        \textbf{Pipe bend} & 9.99 & 9.97 & 0.38 & 9.64 & 10.99 & 10.39 & 10.19 & 0.71 & 9.59 & 11.74 \\
        \textbf{Diffuser} & 30.74 & 30.75 & 0.09 & 30.60 & 30.86 & 30.00 & 30.16 & 0.57 & 29.25 & 30.64\\
        \textbf{Double pipe} & 23.59 &  22.94 & 2.65 & 21.63 & 30.80 & 21.71 & 21.47 & 1.30 & 20.79 & 25.22 \\

        \bottomrule
    \end{tabular}
    \caption{\textbf{Statistics of $\mathcal{J}$ associated with our approach and COMSOL with random initialization:} Our model's statistical metrics are gathered across $10$ training repetitions while  COMSOL's stats are provided for random initializations with $10$ different seed numbers.}
    \label{tab: stats}
\end{table*}
%==========================================================

% We observe that the final dissipated power in the double pipe case is below its analytical value of $32$ reported in \cite{RN1986}. We attribute this common issue to the gray region formed at the pipe boundaries, allowing pipe width to exceed $\frac{1}{6}$. Furthermore, there is some power dissipation in non-fluid regions due to an inadequately large $\alpha_{max}$ which potentially leads to a significant deviation from dissipated power's true value. 

\subsection{Evolution of Topology, Objective Function, and Design Constraint} \label{subsec topology evolution}

\begin{figure*}[!t]
    \centering
    %\vspace{-5pt}
    \begin{subfigure}[b]{1\textwidth}
        \centering
        \includegraphics[width=0.95\columnwidth]{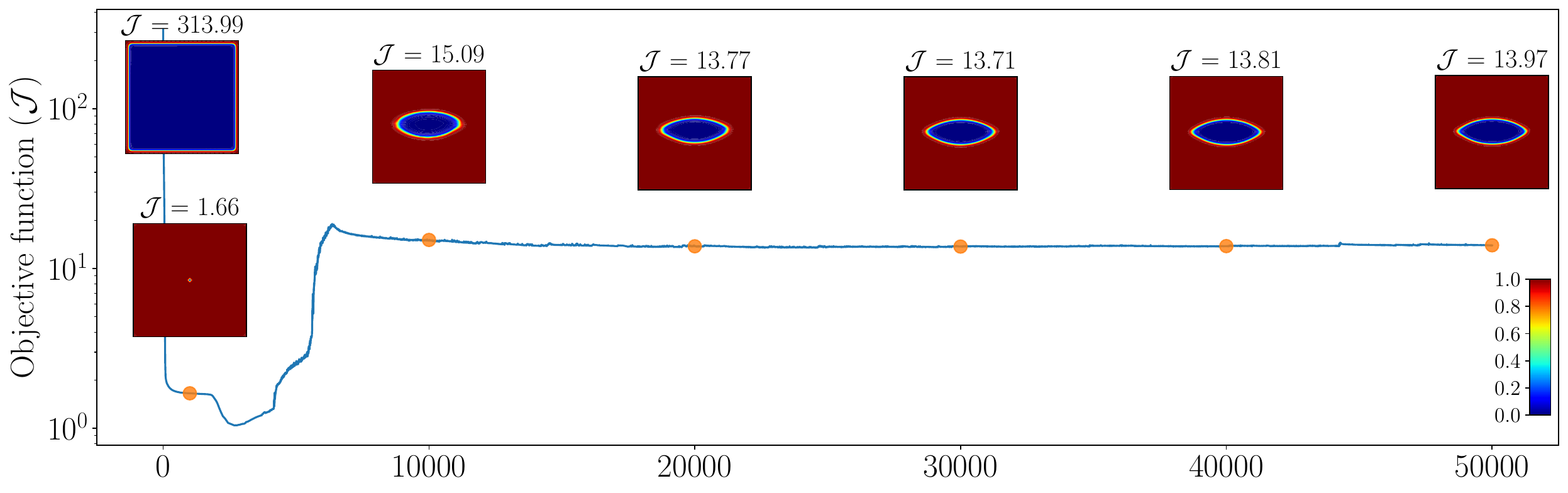}
        %\caption{Rugby BC}
        \label{fig:sub1}
        %\vspace{-5pt}
    \end{subfigure}
    %\vspace{-32pt} % Adjust this value to reduce the vertical space
    \begin{subfigure}[b]{1\textwidth}
        \centering
        \includegraphics[width=0.95\columnwidth]{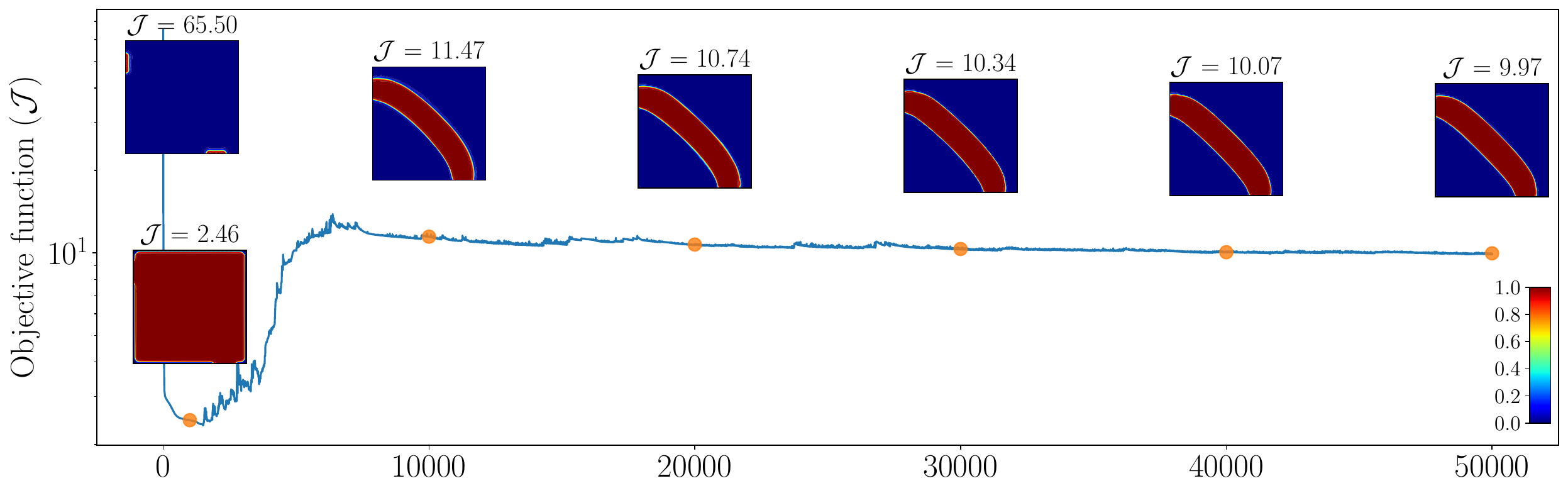}
        %\caption{Pipe bend BC}
        \label{fig:sub2}
        %\vspace{-5pt}
    \end{subfigure}
    % \vspace{-42pt} % Adjust this value to reduce the vertical space
    \begin{subfigure}[b]{1\textwidth}
        \centering
        \includegraphics[width=0.95\columnwidth]{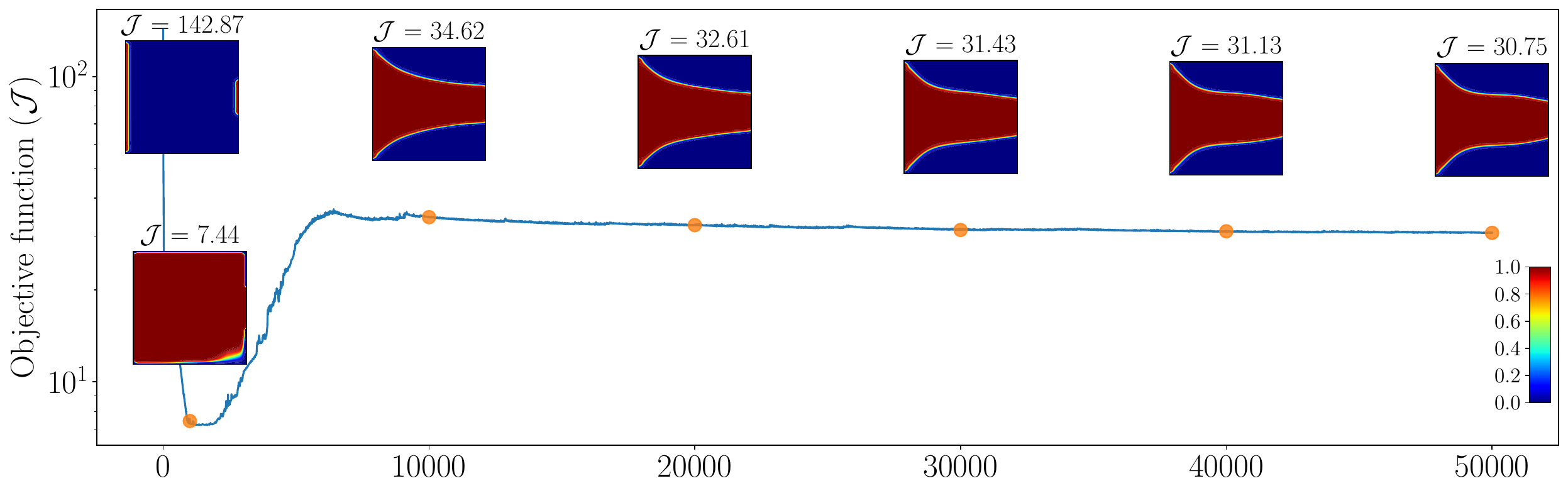}
        %\caption{Diffuser BC}
        \label{fig:sub3}
        %\vspace{-5pt} % Adjust this value to reduce the vertical space
    \end{subfigure}
    
    \begin{subfigure}[b]{1\textwidth}
        \centering
        \includegraphics[width=0.95\columnwidth]{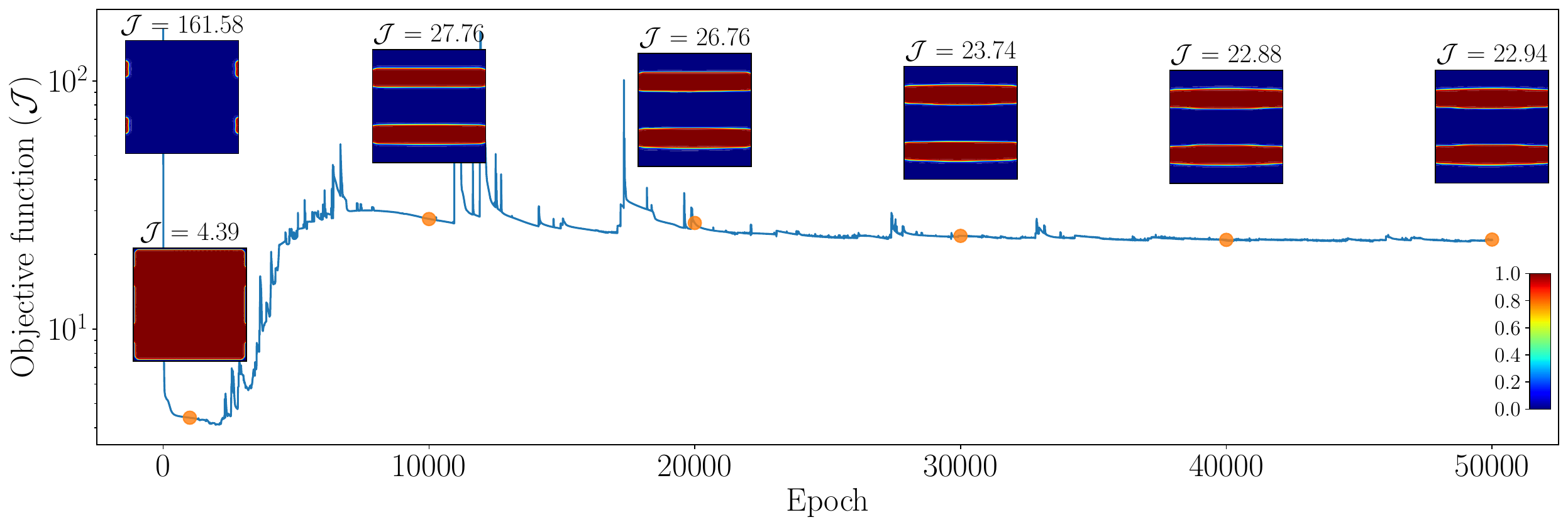}
        %\caption{Double pipe BC}
        \label{fig:sub4}
        \vspace{-5pt}
    \end{subfigure}

    \caption{\textbf{Topology evolution across epochs:} The evolution of $\rho(\nixb)$ is visualized with respect to the objective function over $50,000$ epochs. Optimal topology is obtained as $\mathcal{J}$ converges to its minimum. All quantities are the median across $10$ training repetitions.}
    \label{fig rho evolution}
\end{figure*}

To investigate the training dynamics of our approach, in \Cref{fig rho evolution} we report the topology evolution during the $10$ training repetitions in each problem. Specifically, we visualize the median topology for each problem at epochs $1, 1k, 10k, 20k, \cdots, 50k$ at the corresponding median objective function value. We observe that the $\mathcal{J}$ profiles look quite similar across the four problems where they start at fairly large values, drop very rapidly after about $1k$ training epochs, and then increase relatively quickly to values that are close to the optimum $\mathcal{J}$ in each case. The large fluctuations in the profile of $\mathcal{J}$ roughly disappear at epoch $10k$ after which $\mathcal{J}$ negligibly changes. 

The topology evolution trends in \Cref{fig rho evolution} are similar to those of $\mathcal{J}$ across the four problems in that all the major topological variations take place during the first $10k$ epochs. Specifically, we observe that in all cases after only $1$ epoch $\thetab$ are updated to values that produce $\rho(\nixb) \approx 0$ everywhere in the domain except on the boundaries where the imposed BCs are strictly enforced due to the kernel-weighted corrective residuals, i.e., due to the second term on the right-hand side of \Cref{eq gp conditional mean rho}. After $1k$ of training, the topology is almost flipped in all cases where $\thetab$ take on values that produce $\rho(\nixb) \approx 1$ inside the domain. Both of these two ``extreme" topologies violate the volume fraction constraint but appear during training because they achieve small residuals inside the domain (see also the discussion in \Cref{subsec residuals}). By $10k$ optimization iterations, the topologies in all cases converge and insignificantly change from $10k$ to $50k$ epochs. 

%==================================================

%==================================================

\begin{figure*}[!t]
    \centering
    \includegraphics[width=1.00\columnwidth]{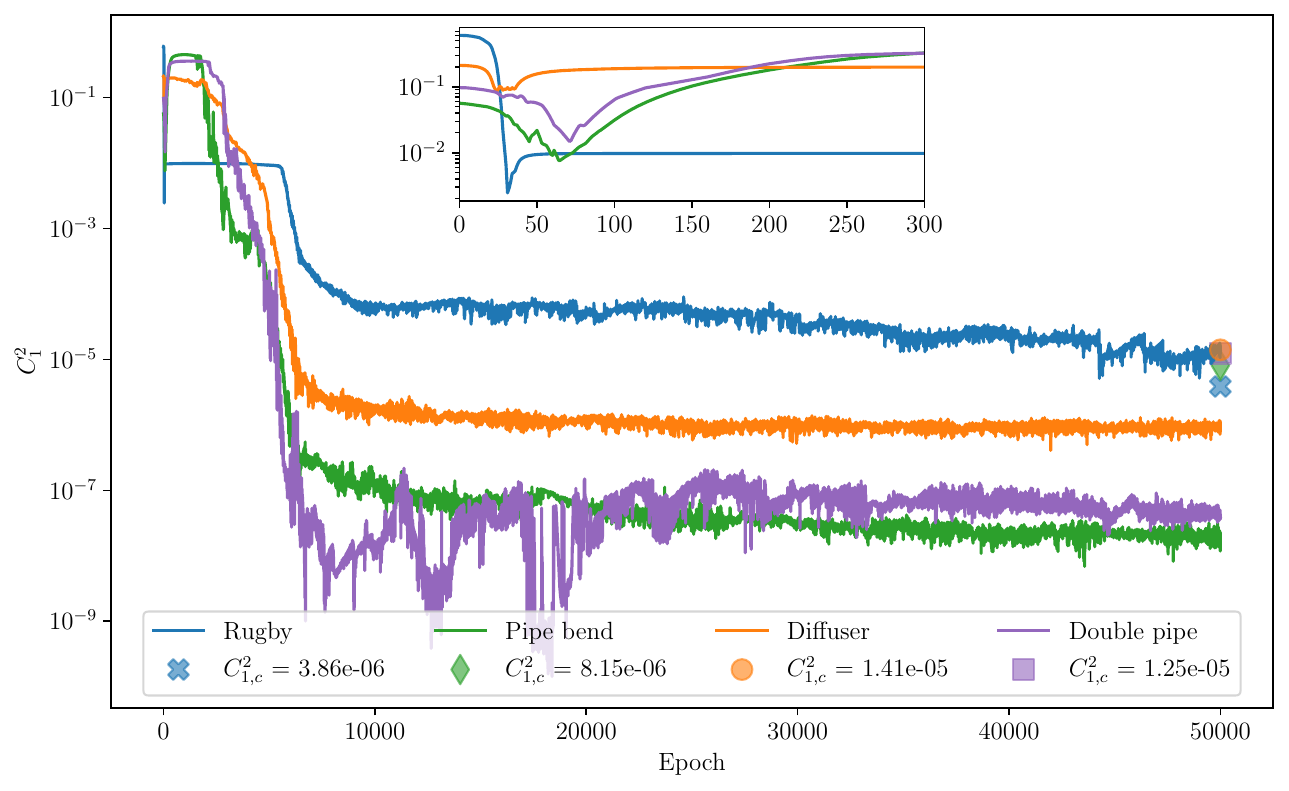}
    \caption{\textbf{Volume loss history:} Each curve represents the median values across $10$ training repetitions. For comparison, in each case we also provide the error obtained by COMSOL (with random initialization) in satisfying the volume fraction constraint, see the markers at epoch $50k$.}
    \label{fig vf evolution}
\end{figure*}

To compare the performance of our approach against COMSOL's in terms of satisfying the design constraint, in \Cref{fig vf evolution} we visualize the profile of the corresponding (unscaled) loss component during the training, i.e., $C_1^2$ in \Cref{eq to dis c1}. We observe that the overall trends are quite similar to those of $\mathcal{J}$ profiles in \Cref{fig rho evolution} where there are some fluctuations during the first few thousands of epochs but then the curves all plateau on average (there are still some fluctuations especially in the case of Double pipe). We notice that the converged values in all cases, except for Rugby, are $1$ to $2$ orders of magnitude smaller than what COMSOL achieves, shown by the markers in \Cref{fig vf evolution}. As discussed in \Cref{subsec summary results,subsec residuals}, even though COMSOL satisfies the design constraint slightly better than our approach in the Rugby example, it provides highly non-optimal topologies in many random realizations. 

%=================================================================
%=================================================================

\subsection{Evolution of PDE Residuals} \label{subsec residuals}

\begin{figure*}[!t]
    \centering
    \includegraphics[width=1.00\columnwidth]{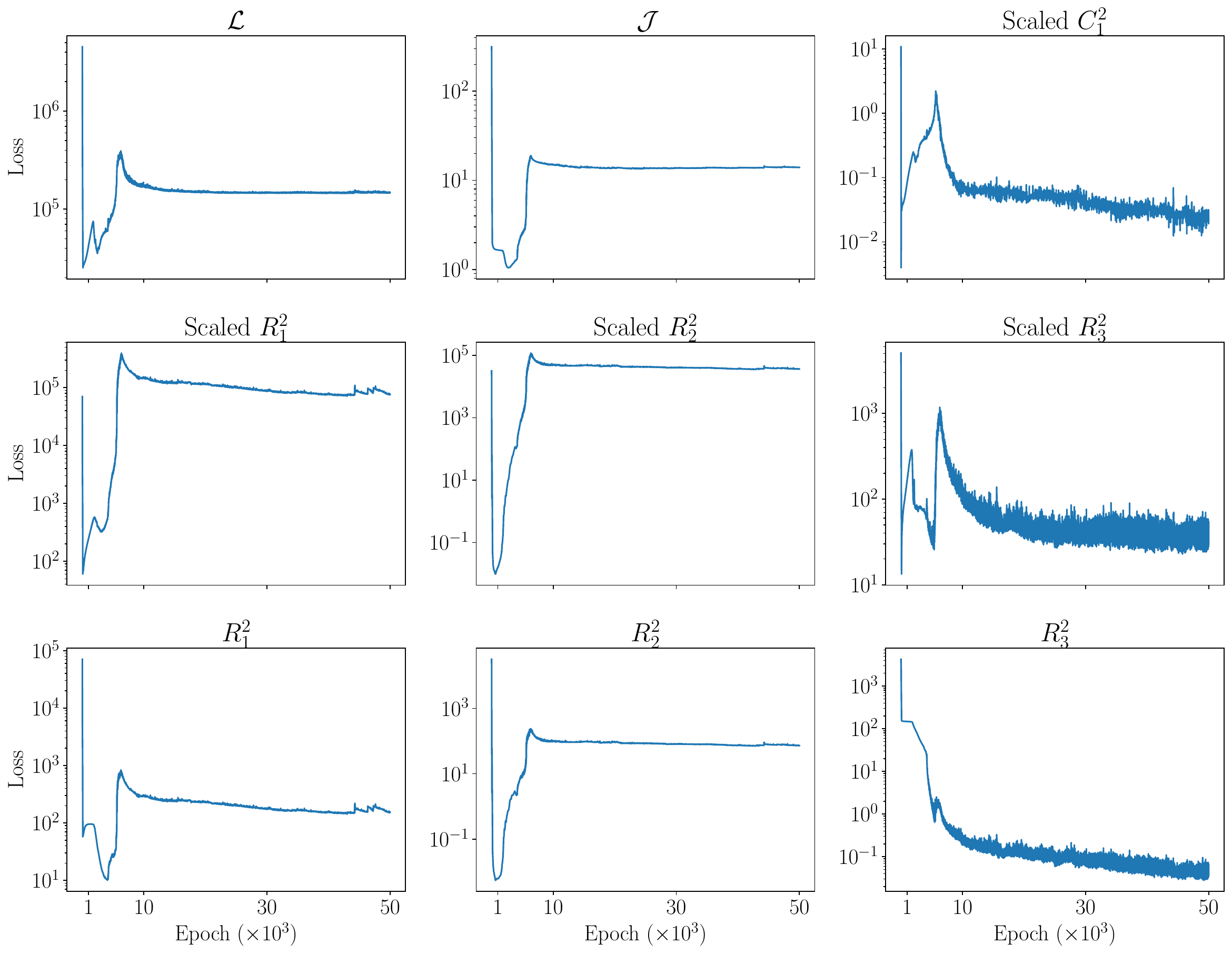}
    \caption{\textbf{Evolution of the loss terms for the Rugby example:} The profile of the individual loss components are quite similar to each other. The unscaled constraint on solid volume fraction is shown in \Cref{fig vf evolution}. Comparison between the scaled and unscaled momentum residuals in the second and third row plots shows the effect of the penalty coefficient in \Cref{eq penalty opt generic 2}. $R_3^2$ corresponds to the continuity equation which is scaled by both the penalty coefficient and the dynamic weight $\alpha$ to ensure its gradient has a similar magnitude to that of the momentum residuals.}
    \label{fig loss components evolution rugby}
\end{figure*}

To analyze the interplay between the loss components, in \Cref{fig loss components evolution rugby} we visualize the history of the individual loss components for the Rugby example (other examples have similar trends, see \Cref{sec more loss curves}). For the state equations (i.e., the continuity and the two momentum equations) both scaled and unscaled versions are provided in the second and third rows of \Cref{fig loss components evolution rugby}, respectively. 
We observe that the profile of the total loss function resembles its five individual components, i.e., $\mathcal{J}$ and the four scaled constraints. We also notice that after about $8k$ optimization iterations, the total loss is largely dominated by the residuals on the two momentum equations which is due to the fact that we use the penalty method to ensure the final design satisfies the state equations. The effect of the penalty coefficient in scaling $R_1^2$ and $R_2^2$ can be observed by comparing the plots in the second and third rows of \Cref{fig loss components evolution rugby}. In the case of the continuity equation, the scaling is due to both the penalty factor and the dynamic weight coefficient in \Cref{eq adam inspired}.

%=================================================================
%=================================================================

\begin{figure*}[!ht]
    \centering
    \includegraphics[width=1.00\columnwidth]{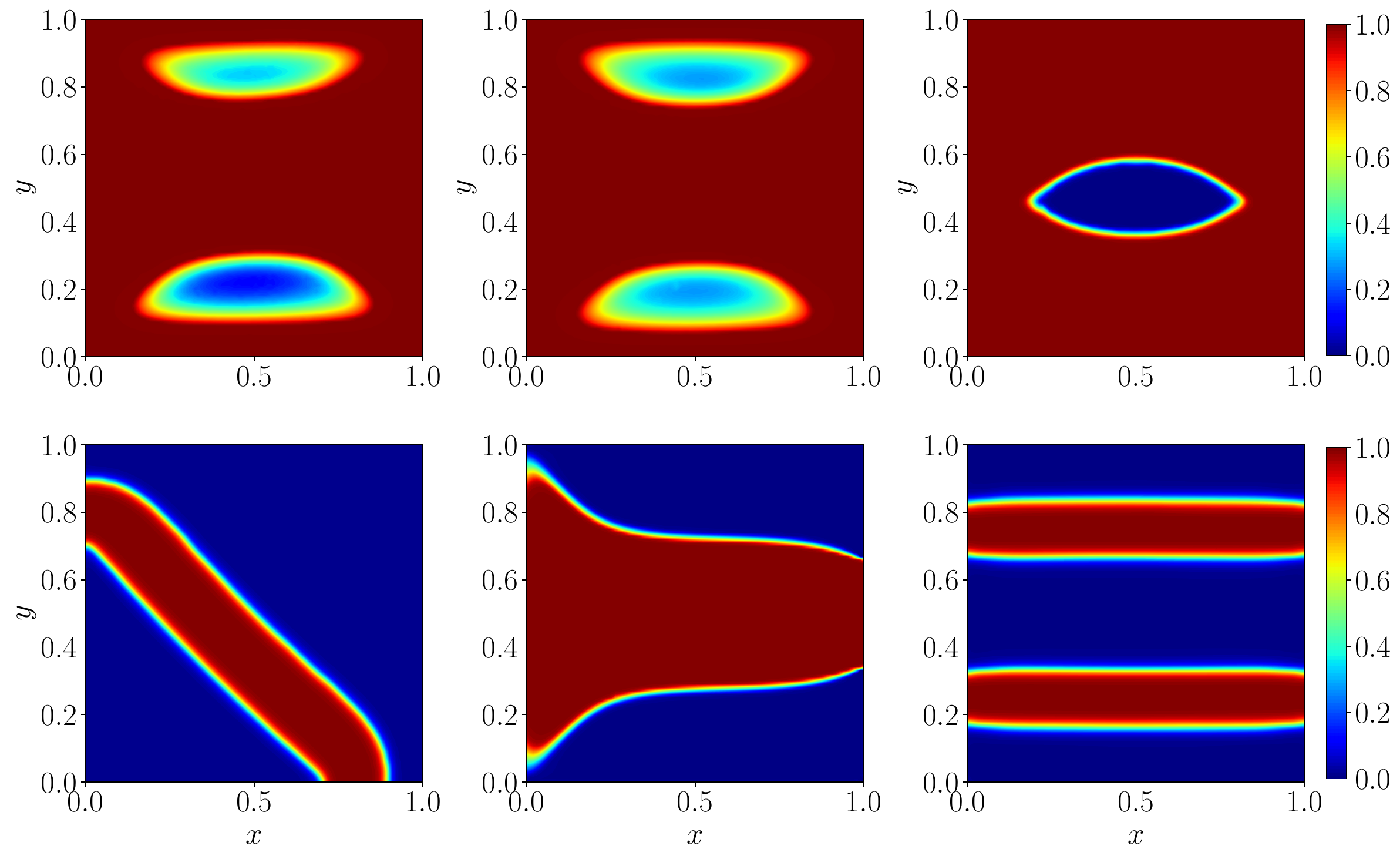}
    \caption{\textbf{Final topologies obtained via COMSOL with random initialization:} Except in the case of Rugby example (top row), the optimum topologies are visually very similar to the ones designed by our approach. The color bar indicates the density which should be ideally $0$ or $1$.}
    \label{fig comsol results}
\end{figure*}

As it can be observed in \Cref{fig loss components evolution rugby} the errors on the three state equations, especially in the case of momentum equations, are fairly large. This observation appears to undermine the quality of our optimum topologies and objective function values. We argue against this statement by noting that the final topologies obtained via our approach are visually identical to those designed by COMSOL if its SIMP-based approach can successfully find the optimum topology, see \Cref{fig comsol results} where COMSOL struggles with the Rugby example. Additionally, as a secondary validation, we import our approach's median optimal topology into COMSOL and solve the state equations in \Cref{eq: Brinkman}. The objective function in \Cref{eq: objective} is then evaluated by COMSOL using a four-point Gaussian quadrature integration over the domain. We denote this objective function by $\mathcal{J}_c$ and report the values for each example in \Cref{tab: results-summary}. It is observed that in all examples $\mathcal{J}_c$ matches very closely to our results.

%=================================================================
%=================================================================

We highlight that the residuals reported in \Cref{fig loss components evolution rugby} are average quantities, i.e., they average the errors evaluated at the CPs which are \textit{uniformly} distributed in the design domain, see \Cref{eq sim to generic c2}. To replace this averaging scheme with one that has a more localized nature, we can $(1)$ weight the residuals at a CP based on the solution gradients at that CP to ensure that sharp transitions at solid-fluid interfaces or inlet-outlet are captured more accurately, or $(2)$ endow the NN mean function with an encoder that parameterizes the design space with learnable features \cite{shishehbor2024parametric}. We do not investigate these extensions in the current work since the obtained topologies and objective function values are better than or competitively close to those of COMSOL. 

\begin{figure*}[!h]
    \centering
    \includegraphics[width=1.00\columnwidth]{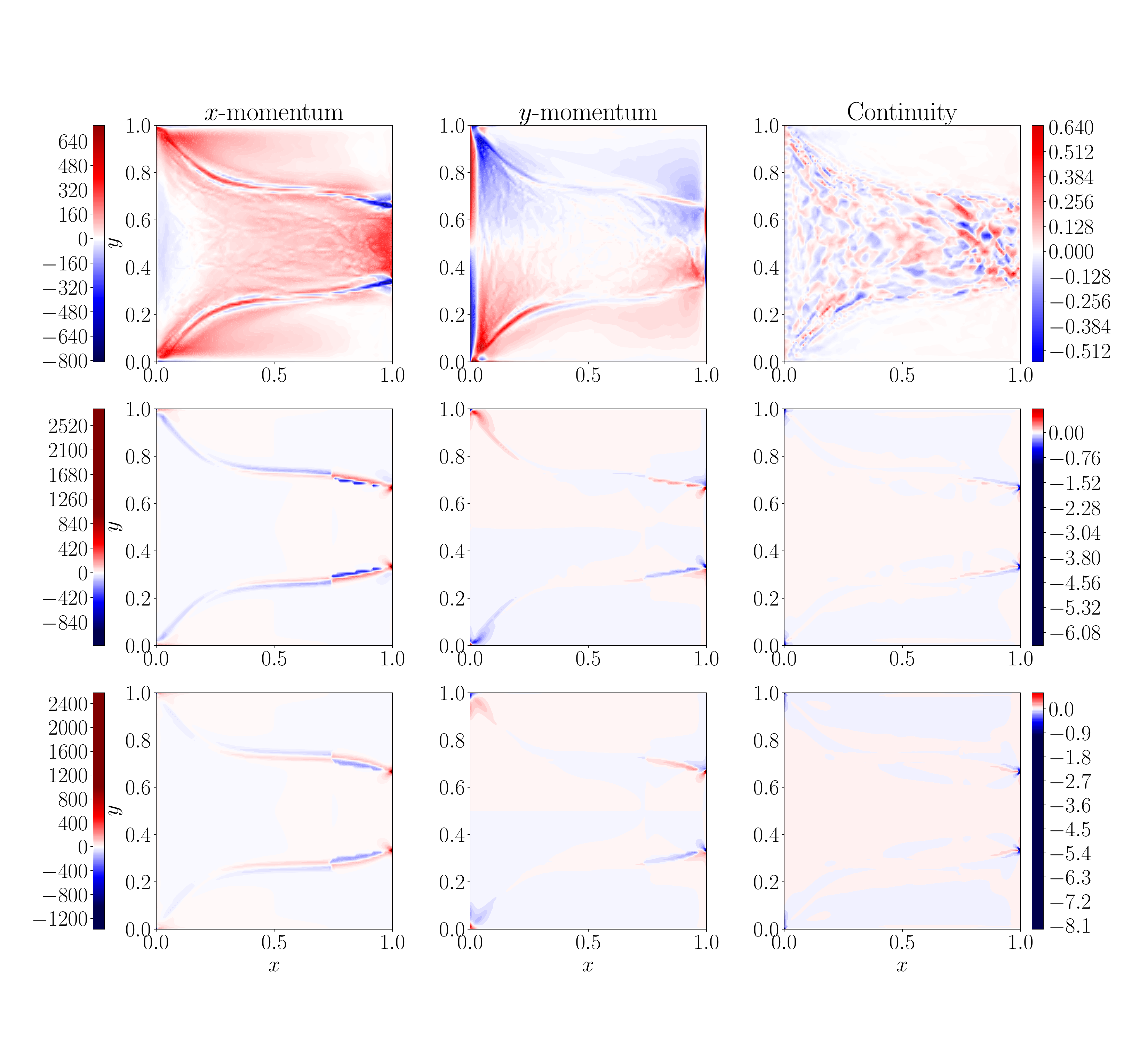}
    \caption{\textbf{Residual maps for Diffuser's median optimal topology:} The top row shows the residuals computed by our model for our median optimal topology. The middle row represents COMSOL's residuals computed for the median optimal topology obtained by our model and the bottom row shows COMSOL's residuals for the median optimal topology designed by COMSOL.  }
    \label{fig: res-map}
\end{figure*}

To visualize the effect of the abovementioned averaging nature and further examine our model's compliance with the underlying physics of the problem, we analyze the Diffuser problem. In \Cref{fig: res-map} we illustrate the residual maps corresponding to the $x$-momentum, $y$-momentum, and continuity equations for our model's median optimal topology and that of COMSOL. In this figure, the top row displays the final median values calculated by the FD scheme of our model, the middle row shows the residuals computed by COMSOL for the median optimal topology obtained by our model, and the bottom row shows the residuals obtained by COMSOL in which the flow is simulated using COMSOL's median optimal topology. The resemblance of the middle and the bottom rows indicates that our model and COMSOL yield very similar optimal Diffuser topologies.

Since our model lacks a localized learning mechanism we expectedly observe in the top row of \Cref{fig: res-map} that the residuals are relatively large in most of the domain, especially near the inlet and outlet where the contributions of the kernel-weighted corrective residuals to the left-hand side of \Cref{eq gp conditional mean} drop and the solution primarily depends on the NN mean function. In contrast, COMSOL's residual maps exhibit smaller values in the majority of the domain for both its own median optimal topology and ours. However, we observe that COMSOL has much higher localized errors at the solid-fluid interfaces and inlet-outlet edges (compare the color bars across the rows). For example, the maximum error incurred in COMSOL for the continuity equation (second and third row) is almost $10$ times larger than the maximum error obtained by our approach (first row).

%=================================================================
%=================================================================

We conclude this section by commenting on the overall shape of the loss profiles shown in \Cref{fig loss components evolution rugby}. The objective function and the four loss terms corresponding to the scaled constraints have two features in common: $(1)$ they have a rough \textit{local} profile, i.e., they fluctuate slightly from one epoch to the next, and $(2)$ their overall trend consists of a sharp drop followed by relatively quick increase to the plateau values. The first feature is natural and expected since each step of the Adam-based optimization process can increase the loss values depending on the parameter updates which are controlled by factors such as the learning rate. The second feature seems slightly counter-intuitive as one expects the total loss to decrease over the course of the optimization but this is not the case in \Cref{fig loss components evolution rugby} for two main reasons: $(1)$ the penalty coefficient which increases gradually until it reaches and stays at its maximum value at around epoch $8k$, and $(2)$ the dynamic weights introduced in \Cref{subsubsec loss weights}. Regarding the latter reason we highlight that $\alpha$ scales a loss term based on that loss term's gradients (and not its magnitude), see \Cref{eq adam inspired}. Hence, loss terms with small magnitudes can still influence the overall trend of the optimization. These terms in our case correspond to the continuity equation and the volume fraction constraint both of which decrease during the optimization (see $R_3^2$ in \Cref{fig loss components evolution rugby} and \Cref{fig vf evolution}) but have smaller \textit{scaled} values than the momentum residuals (compare the magnitude of scaled $C_1^2$ and scaled $R_3^2$ with scaled $R_1^2$ and scaled $R_2^2$ in \Cref{fig loss components evolution rugby}).

% To further demonstrate this point, in \Cref{fig grad hist} we provide the histograms of the gradients of the individual loss components with respect to the parameters of the NN. 

% %=================================================================
% \begin{figure*}[!h]
%     \centering
%     \includegraphics[width=1.00\columnwidth]{figures/grad_hist.pdf}
%     \caption{\textbf{Gradient histograms of the loss terms in the Rugby example:} ...}
%     \label{fig grad hist}
% \end{figure*}
% %=================================================================
    \section {Conclusions and Future Directions} \label{sec conclusion}
We introduce a TO approach based on the framework of GPs whose mean functions are parameterized with deep NNs. Serving as \textit{priors} for the state and design variables, these GPs share a single mean function but have as many independent kernels as there are variables. We develop a computationally robust and efficient scheme for estimating all the parameters of these GP priors within a single loop optimization problem whose objective function is a penalized version of the performance metric (dissipated power in the examples of \Cref{sec results}). The penalty terms account for the state equations as well as design constraints but exclude IC/BCs as these are automatically satisfied by the GP priors. This interpolation feature is not affected upon optimization but the final model is no longer a GP and merely an MAP estimator of the solution.

In \Cref{sec results} we consider four problems where the goal is to identify the topology that minimizes the flow's dissipated power subject to a pre-defined volume constraint. We compare our results against COMSOL which is a commercial software package that leverages SIMP for TO. Our studies demonstrate that our approach $(1)$ is less sensitive to initialization due to its built-in continuation nature, $(2)$ is meshfree and hence dispenses with the filtering techniques that aim to remove mesh-dependency in traditional TO methods, $(3)$ has a consistent computational cost, and $(4)$ is less sensitive to the how the design variable is integrated in the state equations. 

Currently, our approach does not provide a direct mechanism to initialize the topology or control geometrical features; tasks that are quite easily done in traditional methods. A potential direction is to pre-train the mean function on the initial topology and then retrain as detailed in \Cref{sec method}. This idea, however, seems quite costly and ineffective as there is no guarantee that the retraining will preserve the desired features. We believe a more promising direction is to design the architecture of the mean function based on the problem. Such a design can also provide a more localized learning capacity to decrease the residuals on the state equations in regions of the design domain where solution gradients are high (e.g., solid-fluid interface or inlets-outlets). 

In this paper we use maximum number of epochs to terminate the design process but referring to \Cref{fig rho evolution} it is observed that $(1)$ the topology dramatically changes within the first few thousands of epochs but is insignificantly updated after about $8k$ epochs, and $(2)$ the gradient-based optimization problems converge with much fewer epochs. Following these observations, in our future works we plan to develop acceleration schemes and a more systematic method for convergence assessment.

\section*{Acknowledgments}
We appreciate the support from Office of the Naval Research (grant number $N000142312485$) and the National Science Foundation (grant number $2238038$).

    \begin{appendices}

\setcounter{equation}{0}
\renewcommand{\theequation}{A\arabic{equation}}
\setcounter{figure}{0}
\renewcommand\thefigure{A\arabic{figure}}
\setcounter{table}{0}
\renewcommand\thetable{A\arabic{table}}
\renewcommand{\thesection}{A}

\section{Solving PDE Systems with Neural Networks} \label{sec pinns review}
We use the $1D$ viscous Burgers' problem in \Cref{eq burgers} to explain how deep NNs can be used to solve PDE systems:
%=================================================================
\begin{subequations}
    \begin{align}
        &\ut + u \ux - \nu \uxx = 0, && \forall x \in [-1,1], t \in (0, 1] 
        \label{eq burgers pde}\\
        &u(-1, t) = u(1, t) = 0, && \forall t \in [0, 1] 
        \label{eq burgers bc}\\
        &u(x, 0) = -\sin{\parens{\pi x}}, && \forall x \in [-1,1]
        \label{eq burgers ic}
    \end{align}
    \label{eq burgers}
\end{subequations}
%=================================================================
where $x$ and $t$ are the independent variables that denote spatiotemporal coordinates, $u$ is the PDE solution, and $\nu$ denotes the kinematic viscosity.

\begin{figure*}[!hbt]
    \centering
    \includegraphics[width=1.00\columnwidth]{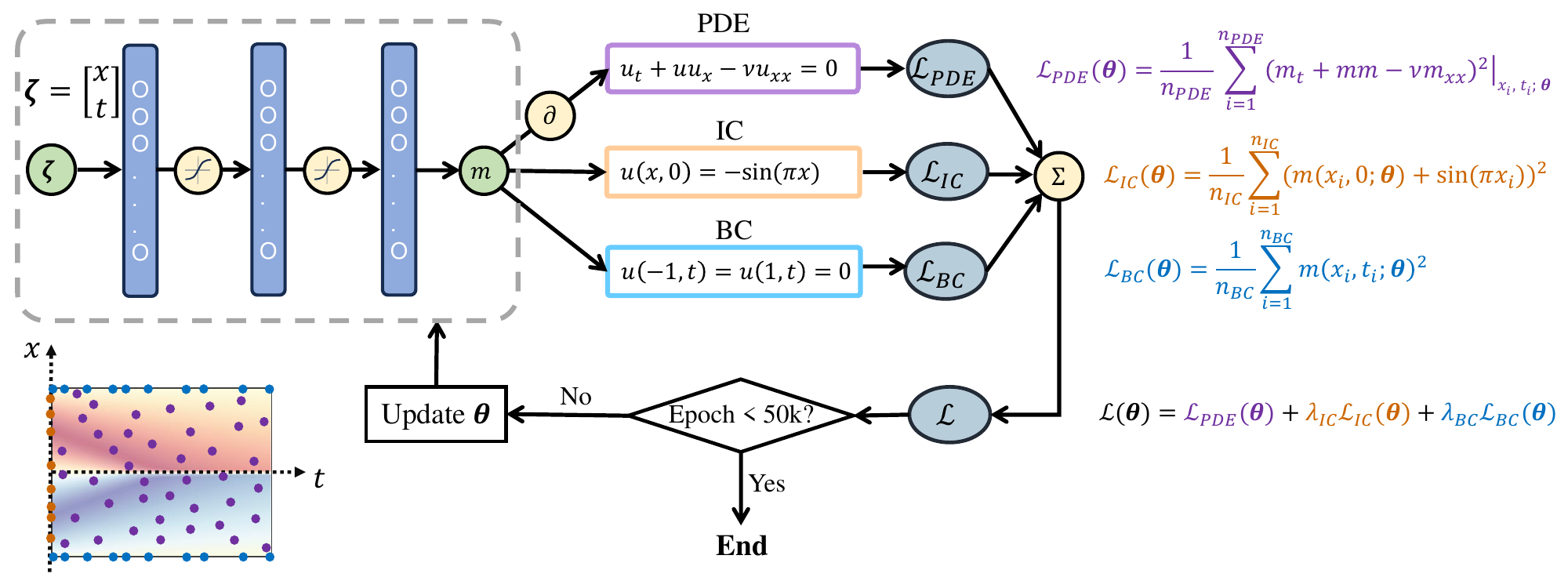}
    \caption{\textbf{Solving the Burgers' problem in \Cref{eq burgers} via a deep neural network:} The model parameters, collectively denoted via $\thetab$, are estimated by minimizing the three-component loss function that encourages the network to satisfy the PDE system in \Cref{eq burgers}. These loss components are calculated by querying the network on a set of test points that are distributed either inside the domain or on its boundaries.}
    \label{fig pinn flowchart}
\end{figure*}

As schematically illustrated in \Cref{fig pinn flowchart}, the main idea of this approach is to parameterize the relation between $u$ and $\brackets{x, t}$ with a deep NN, i.e., $u\parens{x, t} = m(x, t; \thetab)$ where $m(x, t; \thetab)$ denotes the network whose parameters are collectively denoted via $\thetab$.
The parameters of $m(x, t; \thetab)$ are optimized by iteratively minimizing a loss function $\losszeta$ that encourages the network to satisfy the PDE system in \Cref{eq burgers}. 

%=================================================================
%=================================================================

To calculate $\losszeta$, we first obtain the network's output at $n_{BC}$ points on the $x=-1$ and $x=1$ lines, $n_{IC}$ points on the $t=0$ line which marks the initial condition (IC), and $n_{PDE}$ CPs inside the domain. For the $n_{BC} + n_{IC}$ points on the boundaries, we can directly assess the network's predictions based on the BC and IC in \Cref{eq burgers bc,eq burgers ic}. For each of the $n_{PDE}$ CPs, we first evaluate the partial derivatives of the output and then calculate the residual in \Cref{eq burgers pde}. Once these three terms are calculated, we sum them up to obtain $\losszeta$:
%=================================================================
\begin{equation}
    \begin{aligned}
        \losszeta =~ &\mathcal{L}_{PDE}(\thetab) + \lambda_{BC}\mathcal{L}_{BC}(\thetab) + \lambda_{IC}\mathcal{L}_{IC}(\thetab) \\
        =~ &\frac{1}{n_{PDE}}\sum_{i=1}^{n_{PDE}} \parens{m_t(x_i, t_i;\thetab) + m(x_i, t_i;\thetab) m_x(x_i, t_i;\thetab) - \nu m_{xx}(x_i, t_i;\thetab)}^2 +\\
        &\frac{\lambda_{BC}}{n_{BC}}\sum_{i=1}^{n_{BC}} \parens{m(x_i, t_i;\thetab) - 0}^2 
        + \frac{\lambda_{IC}}{n_{IC}}\sum_{i=1}^{n_{IC}} \parens{m(x_i, t_i;\thetab) + \sin{\parens{\pi x_i}}}^2
    \end{aligned}
    \label{eq pinn loss}
\end{equation}
%=================================================================
where $\lambda_{BC}$ and $\lambda_{IC}$ are weights that ensure the scale of the corresponding loss terms is on the same order of magnitude as $\mathcal{L}_{PDE}(\thetab)$. In \Cref{eq pinn loss}, the partial derivatives with respect to $x$ and $t$ are denoted by subscripts. For example, $m_{xx}(\cdot)$ is the second partial derivative of $m(\cdot)$ with respect to $x$.

The loss function in \Cref{eq pinn loss} is typically minimized via either the Adam \cite{kingma2014adam} or the L-BFGS \cite{liu1989limited} methods which are both gradient-based optimization techniques. During the optimization, the parameters of the network are first initialized and then iteratively updated to minimize $\losszeta$. These updates rely on partial derivatives of $\losszeta$ with respect to $\thetab$ which are obtained via automatic differentiation which is also used to calculate the partials $m_x(\cdot), m_t(\cdot),$ and $m_{xx}(\cdot)$.

With either Adam or L-BFGS, it is important to choose appropriate values for $\lambda_{BC}$ and $\lambda_{IC}$ as otherwise $\mathcal{L}_{PDE}(\thetab)$ would dominate the overall loss. These values are typically chosen based on trial-and-error studies and their values change from one PDE system to another. While adaptive methods have been developed to automatically tune these values in case Adam is used as the optimizer, these methods noticeably increase the training costs and fail to generalize to a wide range of PDE systems. 

\setcounter{equation}{0}
\renewcommand{\theequation}{B\arabic{equation}}
\setcounter{figure}{0}
\renewcommand\thefigure{B\arabic{figure}}
\setcounter{table}{0}
\renewcommand\thetable{B\arabic{table}}
\renewcommand{\thesection}{B}

\section{Loss Histories} \label{sec more loss curves}
In the case of the Rugby example, the interplay between the different loss components is discussed in \Cref{subsec residuals}, see \Cref{fig loss components evolution rugby}. The evolution of the loss components for the other three examples look quite similar and are provided below. 

%=================================================================
\begin{figure*}[!t]
    \centering
    \includegraphics[width=1.00\columnwidth]{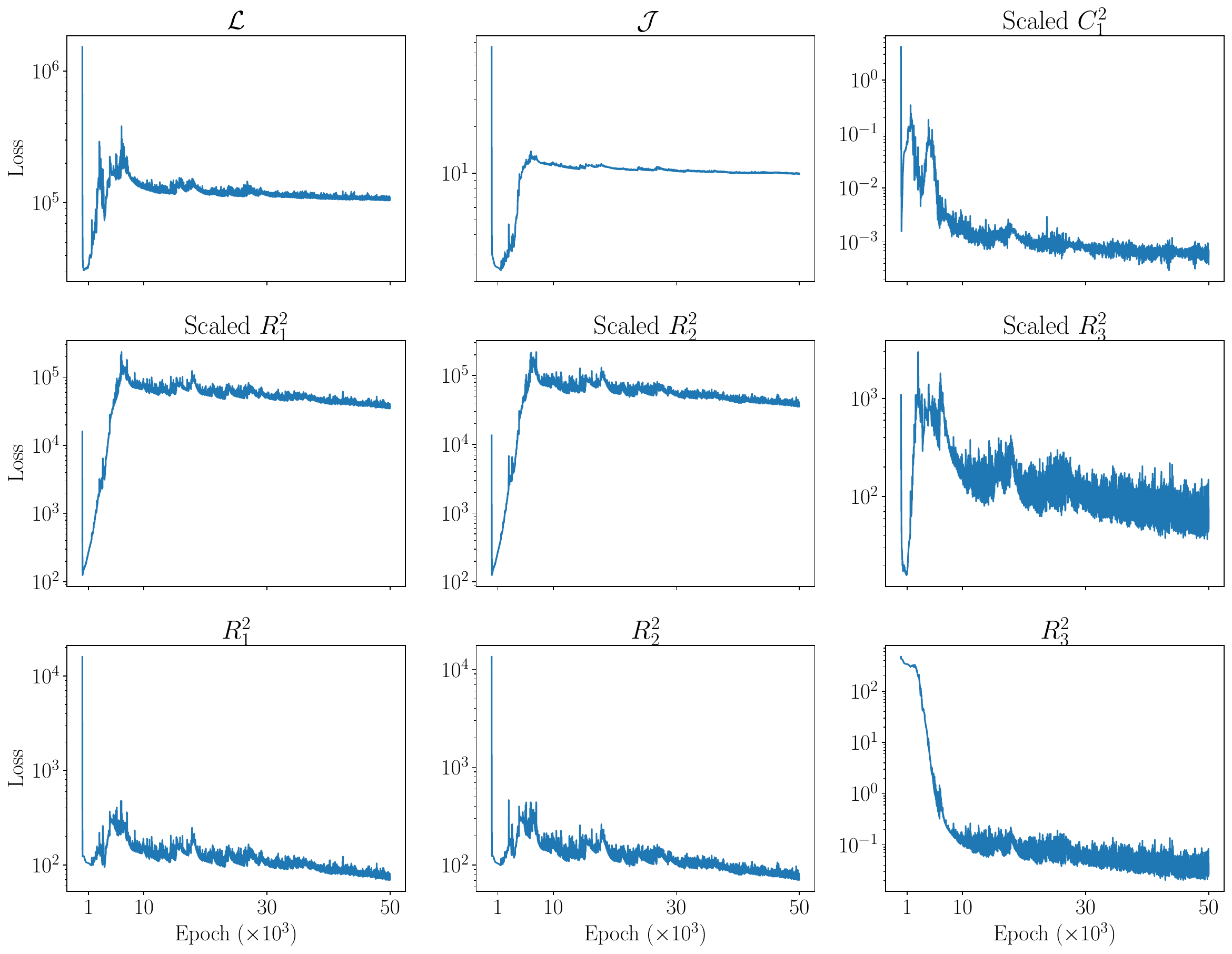}
    \caption{\textbf{Evolution of the loss terms for the Pipe bend example:} The profile of the individual loss components are quite similar to each other and to those in \Cref{fig loss components evolution rugby}. Comparison between the scaled and unscaled momentum residuals in the second and third row plots shows the effect of the penalty coefficient in \Cref{eq penalty opt generic 2}. $R_3^2$ corresponds to the continuity equation which is scaled by both the penalty coefficient and the dynamic weight $\alpha$.}
    \label{fig loss components evolution rugby pipe}
\end{figure*}
%=================================================================

%=================================================================
\begin{figure*}[!t]
    \centering
    \includegraphics[width=1.00\columnwidth]{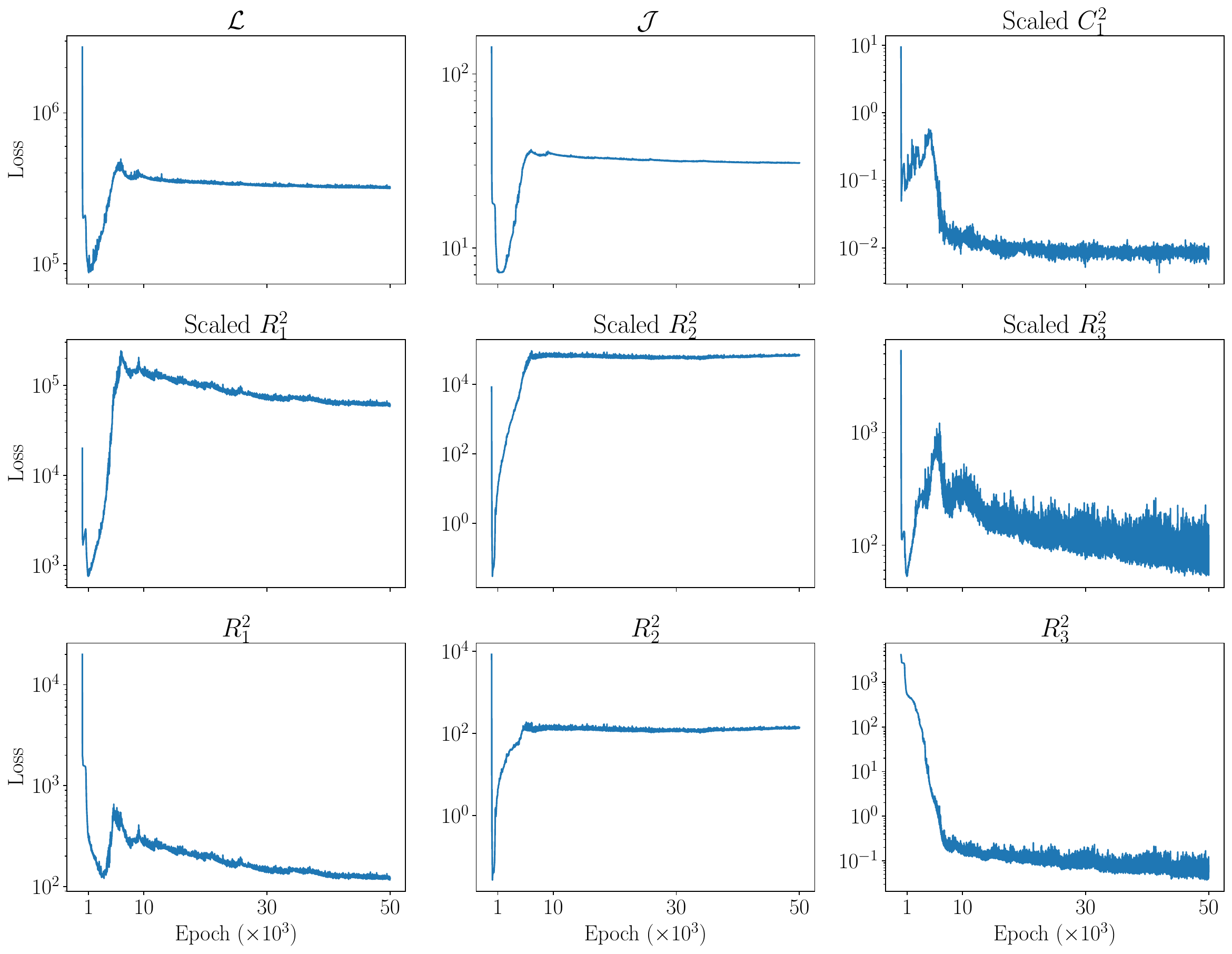}
    \caption{\textbf{Evolution of the loss terms for the Diffuser example:} The profile of the individual loss components are quite similar to each other.}
    \label{fig loss components evolution rugby diffuser}
\end{figure*}
%=================================================================

%=================================================================
\begin{figure*}[!t]
    \centering
    \includegraphics[width=1.00\columnwidth]{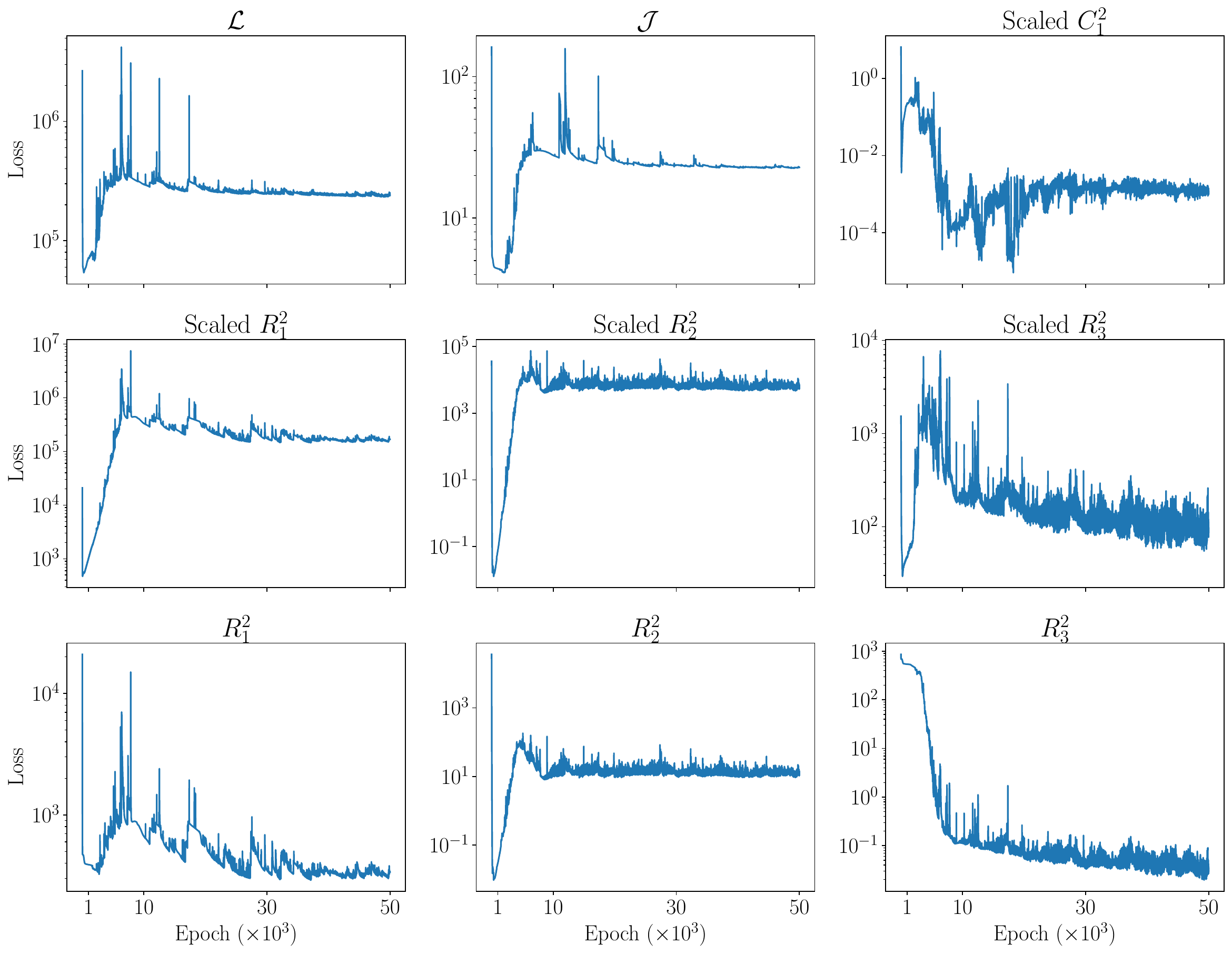}
    \caption{\textbf{Evolution of the loss terms for the Double pipe example:} Compared to the previous examples, the scale of $R_1^2$ at $50k$ epoch is slightly larger while that of $R_2^2$ is one order of magnitude smaller. The profile of the individual loss components are quite similar to each other and to those in other examples.}
    \label{fig loss components evolution rugby double pipe}
\end{figure*}
%=================================================================

\end{appendices}

    \pagebreak
    % Use traditional BibTeX commands
    \bibliographystyle{unsrt} % Choose the appropriate BibTeX style
    \bibliography{01_Ref.bib}     % Use \bibliography to include your .bib file

\begin{thebibliography}{10}

\bibitem{RN2023}
Martin~Philip Bendsøe and Noboru Kikuchi.
\newblock Generating optimal topologies in structural design using a homogenization method.
\newblock {\em Computer methods in applied mechanics and engineering}, 71(2):197--224, 1988.

\bibitem{RN2025}
Martin~P Bendsøe.
\newblock Optimal shape design as a material distribution problem.
\newblock {\em Structural optimization}, 1:193--202, 1989.

\bibitem{RN2050}
Jun Wu, Ole Sigmund, and Jeroen~P Groen.
\newblock Topology optimization of multi-scale structures: a review.
\newblock {\em Structural and Multidisciplinary Optimization}, 63:1455--1480, 2021.

\bibitem{RN2051}
Tomás Zegard and Glaucio~H Paulino.
\newblock Bridging topology optimization and additive manufacturing.
\newblock {\em Structural and Multidisciplinary Optimization}, 53:175--192, 2016.

\bibitem{li2021multidisciplinary}
Shaoying Li, Shangqin Yuan, Jihong Zhu, Weihong Zhang, Han Zhang, and Jiang Li.
\newblock Multidisciplinary topology optimization incorporating process-structure-property-performance relationship of additive manufacturing.
\newblock {\em Structural and Multidisciplinary Optimization}, 63:2141--2157, 2021.

\bibitem{wu2021topology}
Jun Wu, Ole Sigmund, and Jeroen~P Groen.
\newblock Topology optimization of multi-scale structures: a review.
\newblock {\em Structural and Multidisciplinary Optimization}, 63:1455--1480, 2021.

\bibitem{deaton2014survey}
Joshua~D Deaton and Ramana~V Grandhi.
\newblock A survey of structural and multidisciplinary continuum topology optimization: post 2000.
\newblock {\em Structural and Multidisciplinary Optimization}, 49:1--38, 2014.

\bibitem{fin2019structural}
Juliano Fin, Lavinia~Alves Borges, and Eduardo~Alberto Fancello.
\newblock Structural topology optimization under limit analysis.
\newblock {\em Structural and Multidisciplinary Optimization}, 59:1355--1370, 2019.

\bibitem{ben2024robust}
Ismael Ben-Yelun, Ahmet~Oguzhan Yuksel, and Fehmi Cirak.
\newblock Robust topology optimisation of lattice structures with spatially correlated uncertainties.
\newblock {\em Structural and Multidisciplinary Optimization}, 67(2):16, 2024.

\bibitem{stragiotti2024efficient}
Enrico Stragiotti, Fran{\c{c}}ois-Xavier Irisarri, C{\'e}dric Julien, and Joseph Morlier.
\newblock Efficient 3d truss topology optimization for aeronautical structures.
\newblock {\em Structural and Multidisciplinary Optimization}, 67(3):42, 2024.

\bibitem{RN1986}
Thomas Borrvall and Joakim Petersson.
\newblock Topology optimization of fluids in stokes flow.
\newblock {\em International journal for numerical methods in fluids}, 41(1):77--107, 2003.

\bibitem{RN2021}
Michael~Yu Wang, Xiaoming Wang, and Dongming Guo.
\newblock A level set method for structural topology optimization.
\newblock {\em Computer Methods in Applied Mechanics and Engineering}, 192(1):227--246, 2003.

\bibitem{RN2052}
Peng Wei, Zuyu Li, Xueping Li, and Michael~Yu Wang.
\newblock An 88-line matlab code for the parameterized level set method based topology optimization using radial basis functions.
\newblock {\em Structural and Multidisciplinary Optimization}, 58:831--849, 2018.

\bibitem{RN2046}
Zhan Kang and Yiqiang Wang.
\newblock Structural topology optimization based on non-local shepard interpolation of density field.
\newblock {\em Computer methods in applied mechanics and engineering}, 200(49-52):3515--3525, 2011.

\bibitem{RN2048}
Zhen Luo, Nong Zhang, Yu~Wang, and Wei Gao.
\newblock Topology optimization of structures using meshless density variable approximants.
\newblock {\em International Journal for Numerical Methods in Engineering}, 93(4):443--464, 2013.

\bibitem{RN2042}
Seonho Cho and Juho Kwak.
\newblock Topology design optimization of geometrically non-linear structures using meshfree method.
\newblock {\em Computer Methods in Applied Mechanics and Engineering}, 195(44):5909--5925, 2006.

\bibitem{mora2024neural}
Carlos Mora, Amin Yousefpour, Shirin Hosseinmardi, and Ramin Bostanabad.
\newblock Neural networks with kernel-weighted corrective residuals for solving partial differential equations.
\newblock {\em arXiv preprint arXiv:2401.03492}, 2024.

\bibitem{shishehbor2024parametric}
Mehdi Shishehbor, Shirin Hosseinmardi, and Ramin Bostanabad.
\newblock Parametric encoding with attention and convolution mitigate spectral bias of neural partial differential equation solvers.
\newblock {\em Structural and Multidisciplinary Optimization}, 67(7):128, 2024.

\bibitem{RN1920}
Kevin~Stanley McFall and James~Robert Mahan.
\newblock Artificial neural network method for solution of boundary value problems with exact satisfaction of arbitrary boundary conditions.
\newblock {\em IEEE Transactions on Neural Networks}, 20(8):1221--1233, 2009.

\bibitem{RN1187}
Jens Berg and Kaj Nyström.
\newblock A unified deep artificial neural network approach to partial differential equations in complex geometries.
\newblock {\em Neurocomputing}, 317:28--41, 2018.
\newblock Gt1np Times Cited:41 Cited References Count:36.

\bibitem{RN1926}
Justin Sirignano and Konstantinos Spiliopoulos.
\newblock Dgm: A deep learning algorithm for solving partial differential equations.
\newblock {\em Journal of computational physics}, 375:1339--1364, 2018.

\bibitem{RN1886}
Yifan Chen, Bamdad Hosseini, Houman Owhadi, and Andrew~M Stuart.
\newblock Solving and learning nonlinear pdes with gaussian processes.
\newblock {\em Journal of Computational Physics}, 447:110668, 2021.

\bibitem{RN1808}
Nasim Rahaman, Aristide Baratin, Devansh Arpit, Felix Draxler, Min Lin, Fred Hamprecht, Yoshua Bengio, and Aaron Courville.
\newblock On the spectral bias of neural networks, 2019.

\bibitem{RN941}
Diederik~P Kingma and Max Welling.
\newblock Auto-encoding variational bayes.
\newblock {\em arXiv preprint arXiv:1312.6114}, 2013.

\bibitem{RN965}
Ian Goodfellow, Jean Pouget-Abadie, Mehdi Mirza, Bing Xu, David Warde-Farley, Sherjil Ozair, Aaron Courville, and Yoshua Bengio.
\newblock Generative adversarial nets.
\newblock In Z.~Ghahramani, M.~Welling, C.~Cortes, N.~Lawrence, and K.Q. Weinberger, editors, {\em Advances in Neural Information Processing Systems}, volume~27. Curran Associates, Inc., 2014.

\bibitem{li2019non}
Baotong Li, Congjia Huang, Xin Li, Shuai Zheng, and Jun Hong.
\newblock Non-iterative structural topology optimization using deep learning.
\newblock {\em Computer-Aided Design}, 115:172--180, 2019.

\bibitem{RN2027}
Chao Qian and Wenjing Ye.
\newblock Accelerating gradient-based topology optimization design with dual-model artificial neural networks.
\newblock {\em Structural and Multidisciplinary Optimization}, 63(4):1687--1707, 2021.

\bibitem{RN2028}
Heng Chi, Yuyu Zhang, Tsz Ling~Elaine Tang, Lucia Mirabella, Livio Dalloro, Le~Song, and Glaucio~H. Paulino.
\newblock Universal machine learning for topology optimization.
\newblock {\em Computer Methods in Applied Mechanics and Engineering}, 375:112739, 2021.

\bibitem{RN2029}
Vahid Keshavarzzadeh, Robert~M. Kirby, and Akil Narayan.
\newblock Robust topology optimization with low rank approximation using artificial neural networks.
\newblock {\em Computational Mechanics}, 68(6):1297--1323, 2021.

\bibitem{RN1596}
Aaditya Chandrasekhar and Krishnan Suresh.
\newblock Multi-material topology optimization using neural networks.
\newblock {\em Computer-Aided Design}, 136:103017, 2021.

\bibitem{RN2044}
Aaditya Chandrasekhar and Krishnan Suresh.
\newblock Tounn: Topology optimization using neural networks.
\newblock {\em Structural and Multidisciplinary Optimization}, 63(3):1135--1149, 2021.

\bibitem{RN2030}
Seunghye Lee, Hyunjoo Kim, Qui~X. Lieu, and Jaehong Lee.
\newblock Cnn-based image recognition for topology optimization.
\newblock {\em Knowledge-Based Systems}, 198:105887, 2020.

\bibitem{RN2031}
H.~Sasaki and H.~Igarashi.
\newblock Topology optimization accelerated by deep learning.
\newblock {\em IEEE Transactions on Magnetics}, 55(6):1--5, 2019.

\bibitem{RN1024}
M.~Mozaffar, R.~Bostanabad, W.~Chen, K.~Ehmann, J.~Cao, and M.~A. Bessa.
\newblock Deep learning predicts path-dependent plasticity.
\newblock {\em Proc Natl Acad Sci U S A}, 116(52):26414--26420, 2019.

\bibitem{RN1936}
Shiguang Deng, Shirin Hosseinmardi, Libo Wang, Diran Apelian, and Ramin Bostanabad.
\newblock Data-driven physics-constrained recurrent neural networks for multiscale damage modeling of metallic alloys with process-induced porosity.
\newblock {\em Computational Mechanics}, pages 1--31, 2024.

\bibitem{RN1784}
Rebekka~V. Woldseth, Niels Aage, J.~Andreas Bærentzen, and Ole Sigmund.
\newblock On the use of artificial neural networks in topology optimisation.
\newblock {\em Structural and Multidisciplinary Optimization}, 65(10):294, 2022.

\bibitem{RN2034}
B.~S. Lazarov and O.~Sigmund.
\newblock Filters in topology optimization based on helmholtz-type differential equations.
\newblock {\em International Journal for Numerical Methods in Engineering}, 86(6):765--781, 2011.

\bibitem{RN2035}
Atsushi Kawamoto, Tadayoshi Matsumori, Shintaro Yamasaki, Tsuyoshi Nomura, Tsuguo Kondoh, and Shinji Nishiwaki.
\newblock Heaviside projection based topology optimization by a pde-filtered scalar function.
\newblock {\em Structural and Multidisciplinary Optimization}, 44(1):19--24, 2011.

\bibitem{RN2036}
J.~K. Guest, J.~H. Prévost, and T.~Belytschko.
\newblock Achieving minimum length scale in topology optimization using nodal design variables and projection functions.
\newblock {\em International Journal for Numerical Methods in Engineering}, 61(2):238--254, 2004.

\bibitem{RN926}
Ole Sigmund and Kurt Maute.
\newblock Topology optimization approaches.
\newblock {\em Structural and Multidisciplinary Optimization}, 48(6):1031--1055, 2013.

\bibitem{RN2024}
Michael~Yu Wang, Xiaoming Wang, and Dongming Guo.
\newblock A level set method for structural topology optimization.
\newblock {\em Computer Methods in Applied Mechanics and Engineering}, 192(1):227--246, 2003.

\bibitem{RN2037}
Stanley Osher and James~A Sethian.
\newblock Fronts propagating with curvature-dependent speed: Algorithms based on hamilton-jacobi formulations.
\newblock {\em Journal of computational physics}, 79(1):12--49, 1988.

\bibitem{RN2038}
Grégoire Allaire, François Jouve, and Anca-Maria Toader.
\newblock A level-set method for shape optimization.
\newblock {\em Comptes rendus. Mathématique}, 334(12):1125--1130, 2002.

\bibitem{RN2039}
Martin Burger, Benjamin Hackl, and Wolfgang Ring.
\newblock Incorporating topological derivatives into level set methods.
\newblock {\em Journal of Computational Physics}, 194(1):344--362, 2004.

\bibitem{RN2040}
Fabian Wein, Peter~D Dunning, and Julián~A Norato.
\newblock A review on feature-mapping methods for structural optimization.
\newblock {\em Structural and multidisciplinary optimization}, 62(4):1597--1638, 2020.

\bibitem{RN930}
Xiaodong Huang and Yi-Min Xie.
\newblock A further review of eso type methods for topology optimization.
\newblock {\em Structural and Multidisciplinary Optimization}, 41(5):671--683, 2010.

\bibitem{RN2041}
Jonas Zehnder, Yue Li, Stelian Coros, and Bernhard Thomaszewski.
\newblock Ntopo: Mesh-free topology optimization using implicit neural representations.
\newblock {\em Advances in Neural Information Processing Systems}, 34:10368--10381, 2021.

\bibitem{RN2043}
Aditya Joglekar, Hongrui Chen, and Levent~Burak Kara.
\newblock Dmf-tonn: Direct mesh-free topology optimization using neural networks.
\newblock {\em Engineering with Computers}, 2023.

\bibitem{RN2045}
Junyan He, Charul Chadha, Shashank Kushwaha, Seid Koric, Diab Abueidda, and Iwona Jasiuk.
\newblock Deep energy method in topology optimization applications.
\newblock {\em Acta Mechanica}, 234(4):1365--1379, 2023.

\bibitem{RN332}
Carl~Edward Rasmussen.
\newblock {\em Gaussian processes in machine learning}.
\newblock Springer, 2003.

\bibitem{RN1559}
N.~Oune and R.~Bostanabad.
\newblock Latent map gaussian processes for mixed variable metamodeling.
\newblock {\em Computer Methods in Applied Mechanics and Engineering}, 387:114128, 2021.

\bibitem{yousefpour2024gp+}
Amin Yousefpour, Zahra~Zanjani Foumani, Mehdi Shishehbor, Carlos Mora, and Ramin Bostanabad.
\newblock Gp+: a python library for kernel-based learning via gaussian processes.
\newblock {\em Advances in Engineering Software}, 195:103686, 2024.

\bibitem{bertsekas2014constrained}
Dimitri~P Bertsekas.
\newblock {\em Constrained optimization and Lagrange multiplier methods}.
\newblock Academic press, 2014.

\bibitem{wang2021understanding}
Sifan Wang, Yujun Teng, and Paris Perdikaris.
\newblock Understanding and mitigating gradient flow pathologies in physics-informed neural networks.
\newblock {\em SIAM Journal on Scientific Computing}, 43(5):A3055--A3081, 2021.

\bibitem{brinkman1949calculation}
Hendrik~C Brinkman.
\newblock A calculation of the viscous force exerted by a flowing fluid on a dense swarm of particles.
\newblock {\em Flow, Turbulence and Combustion}, 1(1):27--34, 1949.

\bibitem{DUAN20161131}
Xianbao Duan, Feifei Li, and Xinqiang Qin.
\newblock Topology optimization of incompressible navier–stokes problem by level set based adaptive mesh method.
\newblock {\em Computers \& Mathematics with Applications}, 72(4):1131--1141, 2016.

\bibitem{kingma2014adam}
Diederik~P Kingma and Jimmy Ba.
\newblock Adam: A method for stochastic optimization.
\newblock {\em arXiv preprint arXiv:1412.6980}, 2014.

\bibitem{liu1989limited}
Dong~C Liu and Jorge Nocedal.
\newblock On the limited memory bfgs method for large scale optimization.
\newblock {\em Mathematical programming}, 45(1-3):503--528, 1989.

\end{thebibliography}
\end{document}